%% file: 0_ICLR_main.tex
\documentclass{article} % For LaTeX2e
\usepackage{iclr2026_conference,times}

\input{math_commands.tex}

\usepackage{hyperref}
\usepackage{url}

\usepackage{pifont}
\usepackage{cleveref}
\usepackage{graphicx}
\usepackage{booktabs}
\usepackage{multirow}
\usepackage[table]{xcolor}
\usepackage{tcolorbox}
\usepackage{algorithm}
\usepackage{algpseudocode}  % 代替 algorithmic
\usepackage{float}
\usepackage{wrapfig}

\usepackage{titletoc}
\usepackage{enumitem}
\usepackage{caption}
\usepackage{subcaption}

\newcommand{\orangefont}[1]{\textcolor{black}{#1}}
\newcommand{\tealfont}[1]{\textcolor{black}{#1}}

\newcommand{\magentafont}[1]{\textcolor{black}{#1}}
\newcommand{\olivefont}[1]{\textcolor{black}{#1}}
\newcommand{\violetfont}[1]{\textcolor{black}{#1}}

\title{Bridging Global Intent with Local Details: A Hierarchical Representation Approach for Semantic Validation in Text-to-SQL}

\author{
    Rihong Qiu\thanks{Indicates equal contribution.}, \quad \ 
    Zhibang Yang$^{*}$, \quad \ 
    Xinke Jiang$^{*}$ \\ % 这里换一行
    \textbf{
    Weibin Liao, \quad \ 
    Xin Gao, \quad \ 
    Xu Chu, \quad \ 
    Junfeng Zhao, \quad \ 
    Yasha Wang 
    }
    \\
    \\
    Peking University \\
    \texttt{\{rihongqiu,yangzb,xinkejiang\}@stu.pku.edu.cn}
}

\usepackage{enumitem}

\usepackage{xspace}
\newcommand{\methodname}{\textsc{HeroSQL}\xspace}

\iclrfinalcopy % Uncomment for camera-ready version, but NOT for submission.
\begin{document}

\maketitle

\begin{abstract}

Text-to-SQL translates natural language questions into SQL statements grounded in a target database schema. Ensuring the reliability and executability of such systems requires validating generated SQL—but most existing approaches focus only on syntactic correctness, with few addressing semantic validation (detecting misalignments between questions and SQL). As a consequence, how to achieve effective semantic validation still faces two key challenges: capturing both global user intent and SQL structural details, and constructing high-quality fine-grained sub-SQL annotations.
To tackle these, we introduce \methodname, a hierarchical SQL representation approach that integrates global intent (via Logical Plans, LPs) and local details (via Abstract Syntax Trees, ASTs). 
To establish better information propagation, we further employ a Nested Message Passing Neural Network (NMPNN) to capture inherent relational information in SQL and aggregate schema-guided semantics across LPs and ASTs. 
Additionally, to generate high-quality negative samples, we propose an AST-driven sub-SQL augmentation strategy, supporting robust optimization of fine-grained semantic inconsistencies.
Extensive experiments conducted on Text-to-SQL validation benchmarks (in-domain and out-of-domain settings) demonstrate that our approach outperforms existing state-of-the-art (SOTA) methods, achieving an average \olivefont{9.40\%} improvement of AUPRC and \olivefont{12.35\%} of AUROC in identifying semantic inconsistencies.
It excels at detecting fine-grained semantic errors, provides large language models with more granular feedback, and ultimately enhances the reliability and interpretability of data querying platforms. 
% Our code is anonymously available at \url{https://anonymous.4open.science/r/HeroSQL}.

% Text-to-SQL systems, which translate natural language questions into SQL, often generate syntactically correct but semantically flawed queries that fail to capture user intent. While syntactic validation is well-studied, robust semantic validation remains a major challenge due to two primary obstacles: the difficulty of jointly representing global user intent and local SQL structures, and the scarcity of high-quality, fine-grained training data. To address these issues, we introduce HEROSQL, a hierarchical representation framework for SQL validation. HEROSQL integrates global intent using Logical Plans (LPs) with local structural details via Abstract Syntax Trees (ASTs). We employ a Nested Message Passing Neural Network (NMPNN) to effectively propagate and aggregate schema-guided semantic information across this hierarchical structure. Furthermore, to overcome data scarcity, we propose an AST-driven augmentation strategy that generates high-quality, semantically incorrect negative samples for robust model training. Extensive experiments on multiple benchmarks show that HEROSQL significantly outperforms state-of-the-art methods, improving AUPRC by an average of 9.40\% and AUROC by 12.35\% in detecting semantic errors. Our approach not only excels at identifying fine-grained inconsistencies but also provides granular feedback for large language models, thereby enhancing the overall reliability and interpretability of Text-to-SQL systems.

\end{abstract}

\input{1_Introduction}

\input{2_Relatedwork}

\input{3_Preliminaries}

\input{4_Methodology}

\input{5_Experiments}

\input{6_Conclusion}



\bibliography{iclr2026_conference}
\bibliographystyle{iclr2026_conference}

\input{7_Appendix}

\end{document}

%% file: math_commands.tex
\usepackage{amsmath,amsfonts,bm}

\def\eqref#1{equation~\ref{#1}}
\def\1{\bm{1}}

\DeclareMathAlphabet{\mathsfit}{\encodingdefault}{\sfdefault}{m}{sl}
\SetMathAlphabet{\mathsfit}{bold}{\encodingdefault}{\sfdefault}{bx}{n}

%% file: 1_Introduction.tex
\section{Introduction}

% Text-to-SQL validation is very important
Text-to-SQL~\citep{liu2025surveytexttosqlerallms,shi2024survey,10.14778/3681954.3682003} is the task of translating a natural language question into an executable Structured Query Language (SQL) statement, with the translation grounded in the schema of a target relational database.
The fundamental objective of Text-to-SQL is to bridge the ``semantic barrier'' between unstructured user intent (e.g., natural-language queries about data insights) and structured database operations (e.g., filtering, aggregation, and joins)~\citep{liu2025survey}.
Recent research has evolved through three major stages: the \textit{rule-based} stage~\citep{DBLP:journals/corr/abs-2204-00498, 10.1145/2588555.2594519, DBLP:conf/sigmod/Katsogiannis-Meimarakis21, Yu2021Grappa}, the \textit{neural network-based} stage~\citep{xiao2016sequence, lin2019grammar, bogin2019representing}, and the current \textit{pretrained model-based} stage~\citep{li2023graphix, li2022resdsql, gu2023interleaving}. 
More recently, the rapid emergence of Large Language Models (LLMs)~\citep{DBCopilot,cheng2024snil,lian2024chatbi} has further revolutionized Text-to-SQL by providing strong contextual understanding of complex queries and SQL semantics for powerful systems.

\begin{wrapfigure}{r}{9cm}
    \vspace{-0.5cm}
    \raggedright % align to the right
    \includegraphics[scale=0.3]{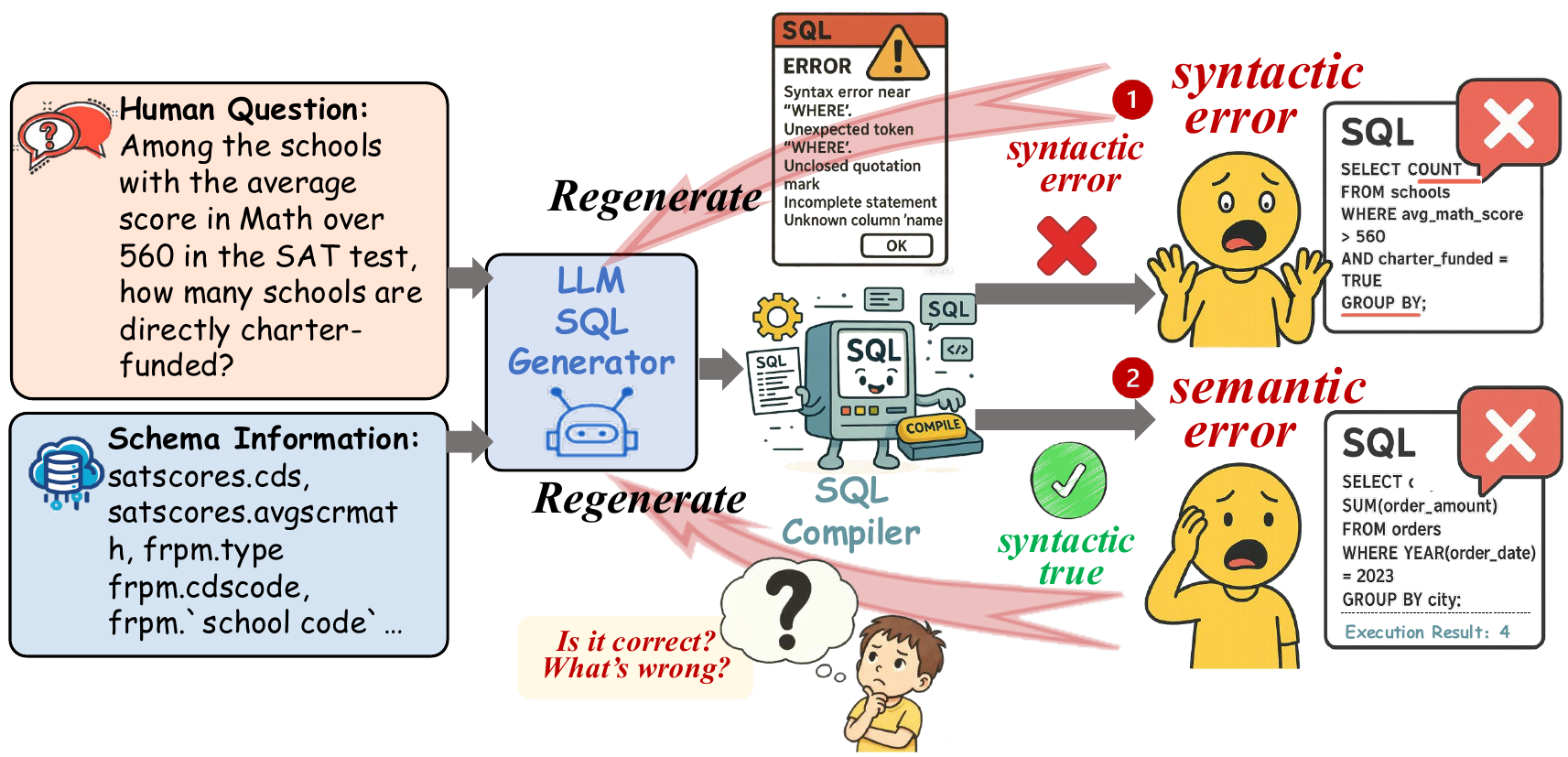}
    \vspace{-0.3cm}
    \caption{Syntactic errors can be easily detected and fixed, while executable SQL with semantic errors is hard to discover.}
    \label{fig:intro}
    \vspace{-0.5cm}
\end{wrapfigure}

Despite significant progress in LLM-based Text-to-SQL systems~\citep{xie2025opensearch,ICLR2025_CHASE_SQL,liao2025learnat,lee-etal-2025-mcs}, these models still frequently generate semantically incorrect queries that may execute successfully but fail to faithfully capture the user’s intent, as shown in Figure~\ref{fig:intro}~\citep{bugfixhard1,bugfixhard2,liu2025nl2sqlbugsbenchmarkdetectingsemantic}.
Unlike \textit{syntactic validation} (e.g., ensuring SQL queries are grammatically correct and executable) with sufficient error feedback, 
\textit{semantic validation} often~\citep{somov2025confidence,askari2025magic,chen-etal-2023-error,chen-etal-2023-sqledit,Arcadinho2022T5QL} aims to identify and correct misalignments between the user’s global intent and the model-generated query structure,  
thereby reducing the cost of human verification and the risk of incorrect results~\citep{bugfixhard1,bugfixhard2}. 
However, 
\textbf{refining semantic validator}~\citep{chen-etal-2023-sqledit} to \textbf{bridge global user intent and local structural database schemas}, 
thereby \textbf{pinpointing more fine-grained misalignment feedback}, motivating several critical research challenges:
% 传统的SQL语义校验的方法难以有效融合全局意图和局部细节信息
% However, existing methods still struggle to effectively integrate both global intent and local details inherent within SQL semantics. 
% Moreover, addressing this issue presents the following challenges.
% 这些不足使我们提出问题：能否找到一个将SQL的整体意图信息和局部结构信息充分融合的高质量模型，以充分表达SQL的内在语义？
% These limitations highlight the need for a model that can effectively combine the global intent with the local structural details of SQL queries, which presents two key challenges \textit{\textbf{C1 \& C2}}.
% thus providing a more comprehensive representation of their underlying semantics.
% These limitations motivate us to ask \textit{is it possible to develop a model that effectively integrates both the global intent and the local structural details of SQL queries?}
% Two challenges for Text-to-SQL validation
%   * [neglect of structure information] capturing the structure information of SQL queries is difficult
%   * [insufficient training data] the training data for Text-to-SQL validation is very limited and the generalization gap

% C1: 如何充分融合SQL query中的全局意图信息和局部结构信息，使得SQL的语义能够被充分表示？和语义信息相结合。之前的xxx工作主要在语义做对齐，近年来有工作考虑了结构信息，但没有充分对齐语义和结构信息
% \noindent\textit{\textbf{\#C1.}}
% \textbf{How to obtain the representation that effectively integrates both the global intent information and the local structural features of the SQL query?} 
\noindent\textit{\textbf{\#C1.}}
\textbf{Challenge in obtaining SQL representations that integrate both local structures and global intent.}
Global intent captures a query’s overarching purpose, encompassing high-level computational logic and data flow. Local structures, by contrast, include schema details, filter/join conditions, and predicate hierarchies that govern fine-grained execution. Integrating these two aspects into a comprehensive semantic representation remains challenging.
% However, existing methods often handle global intent and local structure in isolation, leading to incomplete or fragmented semantic models.

% Thus, it remains a critical research issue to develop strategies that can seamlessly fuse these two levels of information, enabling a more precise and holistic understanding of SQL queries.

% C2: 训练数据量不足，导致模型泛化性不够、能力不足
\noindent\textit{\textbf{\#C2.}}
\textbf{Challenge in acquiring sufficient high-quality and fine-grained training data for Text-to-SQL validation.}
Due to the inherent complexity of the Text-to-SQL task, collecting expert-annotated data that provides fine-grained feedback for iterative SQL refinement is extremely costly—especially at sub-SQL granularity.
This data scarcity significantly limits validator training, preventing models from reliably distinguishing between positive and negative cases at a fine-grained level and ultimately leading to insufficient feedback quality.

% proposed method
To address the preceding challenges, we propose a dual-representation approach for semantic validation in Text-to-SQL.
\ding{182} To address \textbf{\textit{\#C1}}, we design a \textbf{\underline{H}}ierarchical \textbf{\underline{E}}ncoding and \textbf{\underline{R}}epresentation \textbf{\underline{O}}f \textbf{\underline{SQL}} queries (\textbf{\methodname}) for semantic validation. It utilizes the logical plan (LP) to represent the global intent of an SQL query and an abstract syntax tree (AST) to capture local structural details within each node of the logical plan. To fully capture semantic information alongside syntactic structure, we leverage a pretrained embedding model to encode input text with the necessary contextual information (e.g., database schema). We then employ a \emph{Nested Message Passing Neural Network} (NMPNN) to aggregate schema-guided property embeddings, propagating them from local AST nodes to logical plan nodes and further to the entire query plan. 
\ding{183} To address \textbf{\textit{\#C2}}, we propose an \textbf{Adaptive sub-SQL Augmentation Strategy}. Specifically, we introduce semantic perturbations to the AST of gold SQL queries by modifying node attributes within the AST, which enables efficient generation of large-scale syntactically correct yet semantically incorrect negative samples. This enhances the model's ability to distinguish valid queries, particularly at the fine-grained sub-SQL level, improving robustness and generalization. Empirically, we demonstrate that this approach captures fine-grained information, effectively feeding back semantically incorrect sub-SQL for iterative generation, improving overall Text-to-SQL accuracy and significantly reducing human review costs.
% Contributions 
In summary, the main contributions of this work are as follows:
\begin{itemize}[leftmargin=*,noitemsep,topsep=2pt]
    \item We propose \methodname, a hierarchical encoding and representation approach for semantic validation in Text-to-SQL. By leveraging the hierarchical and relational information inherent in the logical plan and abstract syntax trees in SQL query, our approach is capable of capturing both global semantic and local structure information.
    \item We introduce an adaptive sub-SQL augmentation strategy that systematically generates challenging negative examples via AST, effectively mitigating the scarcity of annotated fine-grained SQL validation training data.
    \item Extensive experiments on multiple Text-to-SQL validation benchmarks demonstrate that our approach significantly outperforms existing methods in identifying semantic inconsistencies. It provides more fine-grained feedback on sub-SQL semantic errors and facilitates SQL re-optimization, thereby enhancing the reliability and interpretability of intelligent data analytics platforms.
\end{itemize}

%% file: 2_Relatedwork.tex
\section{Related Work}

\subsection{Semantic Validation in Text-to-SQL}
% While state-of-the-art Text-to-SQL models have made significant progress in translating natural language queries to SQL, they still frequently generate semantically incorrect queries that may execute successfully but produce incorrect results.
% 将SQL直接作为纯文本编码的方法忽略了SQL内在的局部结构信息，难以发现SQL语句中的细节错误
Traditional semantic validation methods for Text-to-SQL~\citep{rai-etal-2023-improving, scholak2021picard, lin2020bridging} treat SQL queries as plain text, employing sequence-encoding models (e.g., BERT~\citep{devlin2019bert}, T5~\citep{raffel2020exploring}) for encoding. These approaches overlook the inherent local structural information of SQL, \textbf{making it difficult to identify subtle errors within statements}.
% 最近，一些研究工作开始考虑融合SQL语句中的结构信息
Recently, some studies have begun incorporating structural information from questions and SQL queries~\citep{Feng2020CodeBERT, Yu2021Grappa}. 
Some of these methods model questions and database schemas as interconnected graphs~\citep{Wang2020RATSQLa, hui2022s, qi2022rasat, bazaga2023sqlformer}, while others represent SQL queries using abstract syntax trees (AST)~\citep{chen-etal-2023-error, gong2025sqlens}. 
Unlike plain text, SQL queries have a natural nested structure—their inherent hierarchical structural information is critical for representing SQL's semantic meaning.
For example, SQLens~\citep{gong2025sqlens} predicts clause-level semantic correctness in SQL queries by aggregating weak supervision signals from DB-based checks and LLM-based evaluations on AST.
% 然而，AST更多的是展现SQL语句的语法细节，难以从宏观上捕获SQL语句的全局语义，找出SQL语句的意图错误
Yet these approaches focus largely on the syntactic details of questions and SQL queries, \textbf{making it difficult to capture global semantic intent from a macroscopic perspective}.
% % structure information is important for Text-to-SQL validation
% This syntactic hierarchy directly encodes relationships between different query components, thereby shaping the overall logic and intent of the query beyond the lexical level. 
% Therefore, understanding this structural information is essential for accurately validating the semantic correctness of SQL queries.

\subsection{Data Augmentation for Text-to-SQL}
% 由于Text-to-SQL的复杂性，在这个领域的训练数据非常稀缺
Due to the intrinsic complexity of the Text-to-SQL task, assembling large-scale, high-quality annotated datasets is extremely challenging. 
% 近年来，合成数据被证明是有效的
To mitigate this, numerous data augmentation strategies have been proposed to generate synthetic data, which has been proven effective in improving model generalization for Text-to-SQL~\citep{hu2023importance,tarbell2024towards}.
% Text-to-SQL数据构造方法，rule-based
Early approaches primarily relied on rule-based and template-driven methods for data augmentation~\citep{lee-etal-2025-mcs, li2024codes, ScienceBenchmark}, which \textbf{limit data diversity and domain coverage}.
% llm-based
Recent approaches have turned to LLMs to generate questions and corresponding SQL queries~\citep{duan2025dsqg, OmniSQL, 10484107, yang-etal-2024-synthesizing}, 
enabling the creation of more diverse and domain-specific training samples. 
However, \textbf{the computational and financial costs associated with employing LLMs remain prohibitively high}.
% 注意到CodeBERT中提出了AST构造负样本训练数据方法，但是没有用到Text-to-SQL里面
Notably, some methods, such as CodeBERT~\citep{Feng2020CodeBERT} in the code generation domain, attempt to construct negative samples using abstract syntax trees, which enables more structured and semantically meaningful examples for model training.
However, \textbf{such techniques have not been explored within the Text-to-SQL domain}, presenting a promising avenue for generating large quantities of negative samples in a cost-effective manner.

%% file: 3_Preliminaries.tex
\section{Preliminaries}
% In Text-to-SQL tasks, errors can be broadly categorized into two categories, which are \textit{syntax errors} and \textit{semantic errors}~\citep{liu2025nl2sqlbugsbenchmarkdetectingsemantic}. 
% \textbf{\textit{Syntax errors}} occur when an SQL query violates the rules of SQL grammar or database-specific constraints. These errors are typically easy to detect, since the database engine returns explicit error messages during query execution.
% In contrast, \textbf{\textit{Semantic errors}} refer to queries that are syntactically correct but do not capture the intended meaning of the user's natural language question. Such errors are much harder to identify automatically, as SQL queries can be executed successfully without returning an error, but the results do not match user expectations.
% Therefore, our validation task focuses on the detection of semantic errors in Text-to-SQL tasks.

\paragraph{\textit{Task Definition 1. (Text-to-SQL Validation)}}
The Text-to-SQL task aims to translate a natural language question $q$ into a corresponding SQL query $s$ that is executable on a target relational database with a predefined schema $\texttt{SCHEMA}$. As a critical post-hoc component of the Text-to-SQL pipeline, Text-to-SQL validation focuses on assessing the correctness of the generated SQL query $s$ with respect to both the input question $q$ and the database schema $\texttt{SCHEMA}$. The probability score $\hat{y}$ for the prediction can thus be formalized as Equation~\ref{eq:task-definition}, where $f$ denotes the validation function:
\begin{equation}
\label{eq:task-definition}
  \hat{y} = f(q, s \mid \texttt{SCHEMA}),
\end{equation}
where $s$ can either be a complete SQL query or a sub-SQL fragment derived from a Logical Plan (LP). For sub-SQL fragments, we can utilize validation scores to deliver fine-grained error feedback to support LLMs in optimizing subtle semantic errors that are otherwise difficult to detect.

\paragraph{\textit{Definition 2. (Intermediate Representations of SQL Queries)}} 
For structured queries such as SQL, it is crucial to construct \textit{intermediate representations} (IRs) that capture both global semantics and local syntactic details. Two widely adopted IRs are the \textit{logical plan} (LP) and the \textit{abstract syntax tree} (AST), which provide complementary perspectives: LPs encode the high-level semantic intent of query execution, while ASTs preserve the fine-grained syntactic organization of SQL statements.

\begin{itemize}[leftmargin=*]
    \item \textbf{\ding{182} Logical Plan (LP).} 
    A Logical Plan is a kind of semantic tree that offers a structured, high-level abstraction of a database query, formalizing the core sequence of operations required to derive the intended result. 
    A logical plan abstracts a query as a structured sequence of relational operations (e.g., \texttt{Filter}, \texttt{Join}, \texttt{Aggregate}) that collectively define how the query result is derived. 
    Intuitively, an LP can be represented as a directed acyclic graph 
    $\mathcal{G}_{\mathcal{LP}} = (V_{\mathcal{LP}}, E_{\mathcal{LP}})$, 
    where $V_{\mathcal{LP}} = \{(o_i, a_i)\}$ includes a set of operators that each operator $o_i$ is associated with attributes $a_i$ that describe its properties, and $E_{\mathcal{LP}}$ encodes the flow of intermediate results between operators. 
    % \begin{equation}
    %     \mathcal{LP} = (o_1 \!\to\! o_2 \!\to\! \cdots \!\to\! o_n),
    % \end{equation}
    % Intuitively, an LP can be expressed as an operator 
    % \begin{equation}
    %     \mathcal{LP} = (o_1 \!\to\! o_2 \!\to\! \cdots \!\to\! o_n),
    % \end{equation}
    % where each operator $o_i$ may have associated attributes $a_i$ describing its properties. More generally, the LP can be represented as a directed acyclic graph 
    This representation abstracts away low-level execution details while \textbf{capturing the global semantic flow of the query}.

    \item \textbf{\ding{183} Abstract Syntax Tree (AST).} 
    In contrast, an AST captures the detailed syntactic structure of an SQL query as a hierarchical, tree-shaped form. Each node corresponds to a syntactic construct such as \texttt{SELECT}, \texttt{FROM}, \texttt{WHERE} clauses, tables, columns, or predicates. 
    Formally, an AST is denoted as $\mathcal{AST}$ and represented as 
    $\mathcal{G}_{\mathcal{AST}} = (V_{\mathcal{AST}}, E_{\mathcal{AST}})$, 
    where each node $v_j^{\mathcal{A}} \in V_{\mathcal{AST}}$ contains atomic syntactic content $c_j^{\mathcal{A}}$, and edges $E_{\mathcal{AST}}$ capture the hierarchical nesting among components. 
    By removing superficial syntax (e.g., parentheses, redundant keywords), ASTs provide a normalized view emphasizing local syntactic relationships and hierarchical structure. This granular, node-based structure \textbf{enables fine-grained analysis of local syntactic patterns and supports precise detection of subtle structural inconsistencies} in SQL statements.
\end{itemize}

% By integrating LPs and ASTs, we obtain a hierarchical intermediate representation that simultaneously encodes the query's global intent and local syntactic details, enabling precise semantic and structural reasoning over SQL statements.

%% file: 4_Methodology.tex
\section{Methodology}

% In this section, 

% 训练数据增强和训练算法
    % 人标数据
        % 原始数据集的ground truth
    % 机器标数据怎么来
        % LLM生成了一批正负样本
        % 语义结构负样本增强
            % 基于AST扰动生成一批负样本
            % 基于LP扰动生成一批负样本
    % 训练策略
        % SFT + 对比学习
        % 加速策略
 
% 4.1 比如C1就是Global和Local的结合：

% subsection 4.1.1 结构构造：就是如何构造AST和LP的双层结构
% subsection 4.1.2 前缀引导+大模型embedding模型做编码，获得语义信息
% subsection 4.1.3 双层message passing NNs

% ——————

% 然后4.2 就是针对数据不够的策略
% subsection 4.2.1 就是ground truth数据介绍
% subsection 4.2.2 就是弱监督数据怎么来
%     paragraph 1: 用大模型生成
%     paragraph 2: 结构扰动策略
% subsection 4.2.3 就是训练策略--加入SFT和对比学习

% ————————

% 然后4.3 讲清楚现实需求，因为速度要求,所以引入了一些加速策略  
 
\begin{figure}[t!]
    \centering
    \includegraphics[width = 1.0\linewidth]{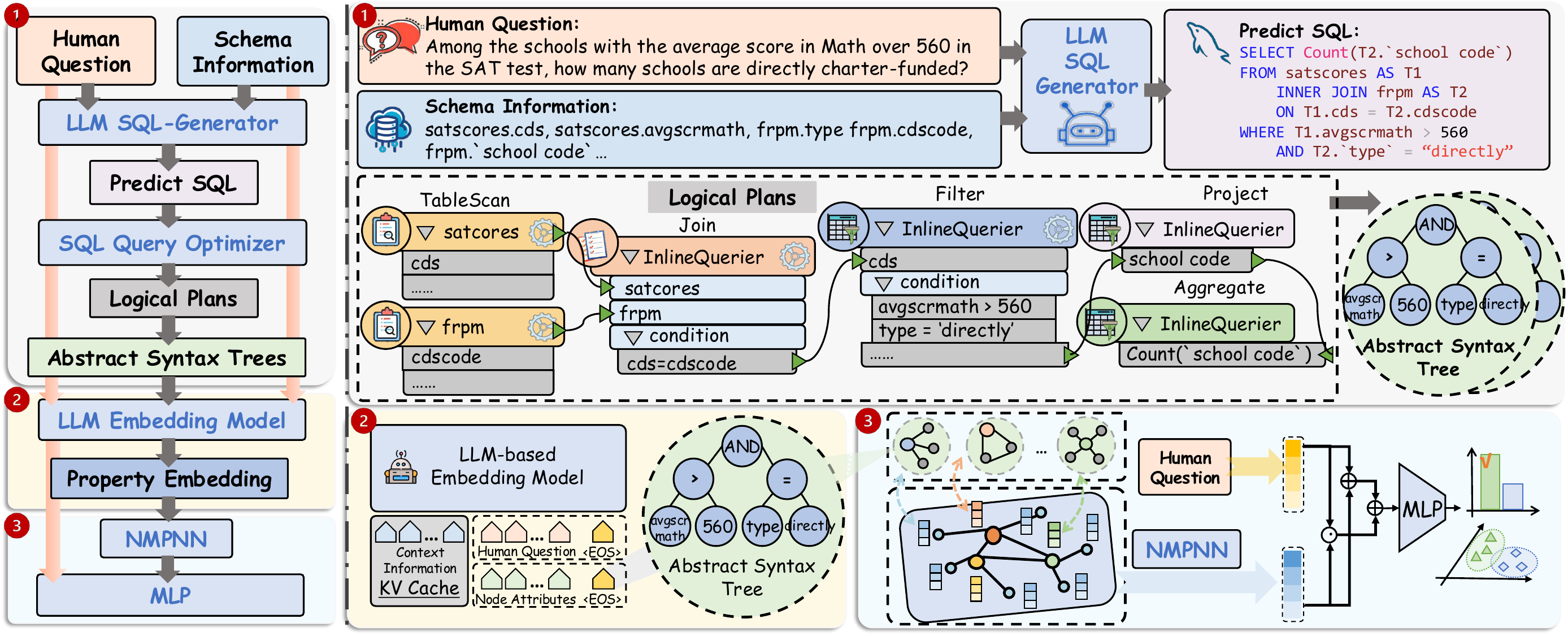}
    \caption{
        The overall framework of \methodname.
        Given a natural language question, database schema information, and a syntactically correct SQL query generated by an LLM, 
        \methodname first applies a query optimizer to convert the query into a logical plan (LP) and parses each LP node into an abstract syntax tree (AST). The AST nodes and contextual information are then encoded into property embeddings using a pretrained LLM-based model. 
        Finally, these embeddings are aggregated by NMPNN to form a SQL embedding, which, together with the question embedding, is fed into an MLP for semantic validation and correctness prediction.
    }
    \label{fig:framework}
\end{figure}

% TODO: 和下面内容有一点重复
In this section, we present the methodology of \methodname, as illustrated in Figure~\ref{fig:framework}. 
First, in Section~\ref{Hierarchical Encoding and Representation of SQL Queries}, we describe how SQL queries are represented and encoded through a hierarchical structure that integrates both logical plans and abstract syntax trees. 
Next, in Section~\ref{Data Augmentation for Text-to-SQL Validation}, we introduce our data augmentation strategies, which enrich the training corpus with diverse and fine-grained sub-SQLs for robust validation. 
Finally, in Section~\ref{Efficient Training of SQL Validation Model}, we detail the training procedure of our SQL validation model and discuss several optimization and acceleration techniques that enhance both efficiency and scalability.

% a hierarchical representation approach of SQL queries that integrates both global intent and local details of SQL queries with logical plan (LP) and abstract syntax trees (AST). 

% TODO: 图加一点注意力引导，现在有点乱

\subsection{Hierarchical Encoding and Representation of SQL Queries}
\label{Hierarchical Encoding and Representation of SQL Queries}

In Section~\ref{sec:dual-representation}, we first introduce the hierarchical representation approach that effectively represents global semantics and local structure details of the SQL query with a logical plan and an abstract syntax tree. 
Then, we describe the embedding strategy for the text property in the context with schema information and the complete SQL query in Section~\ref{sec:schema-guided-embedding}. 
Finally, we present the \emph{Nested Message Passing Neural Network} that aggregates all information for both global semantics and local structure in the hierarchical representation of SQL queries in Section~\ref{sec:nested-MPNN}.

\subsubsection{Hierarchical Intermediate Representation Construction}
\label{sec:dual-representation}

% Benefiting from the complementary strengths of logical plans (LPs) in capturing high-level semantics and abstract syntax trees (ASTs) in modeling fine-grained structural details, we construct a \textit{hierarchical intermediate representation} (HIR) for SQL queries. 
We construct a hierarchical intermediate representation (HIR) for SQL queries, which benefits from the complementary strengths of logical plans (LPs) for capturing high-level semantics and abstract syntax trees (ASTs) for modeling fine-grained structural details.
This representation jointly encodes global semantic intent and local structural information.

\paragraph{\ding{182} Global Intent via Logical Plans.}  
To model the global intent of a SQL query, we employ query optimizers such as Apache Calcite~\citep{begoli2018apache} and ORCA~\citep{soliman2014orca} to convert the raw SQL query into its corresponding logical plan, represented as a directed acyclic graph:
\(
\mathcal{G}_{\mathcal{LP}} = (V_{\mathcal{LP}}, E_{\mathcal{LP}}),
\)
where each node $v_i^{\mathcal{LP}} = (o_i, a_i) \in V_{\mathcal{LP}}$ consists of a relational operator $o_i$ (e.g., \texttt{Filter}, \texttt{Join}, \texttt{Aggregate}) and its associated attribute $a_i$. The edges $E_{\mathcal{LP}}$ encode the flow of intermediate results between operators, capturing high-level semantic dependencies of the query.

\paragraph{\ding{183} Local Structure via Abstract Syntax Trees.}  
After obtaining each LP node  (sub-SQL) $v_i^{\mathcal{LP}}$, the attribute $a_i$ (e.g., a filter condition or an aggregate expression) is transformed into an abstract syntax tree $\mathcal{A}_i$ to capture fine-grained syntactic structure:
\(
\mathcal{G}_{\mathcal{A}_i} = (V_{\mathcal{A}_i}, E_{\mathcal{A}_i}),
\)
where each node $v_j^{\mathcal{A}_i} \in V_{\mathcal{A}_i}$ contains atomic syntactic content $c_j^{\mathcal{A}_i}$, and edges $E_{\mathcal{A}_i}$ represent hierarchical composition and operator precedence. 

\paragraph{\ding{184} Hierarchical Construction.}  
By combining the LP $\mathcal{G}_{\mathcal{LP}}$ with its node-specific ASTs $\{\mathcal{A}_i\}_{v_i \in V_{\mathcal{LP}}}$, we obtain hierarchical intermediate representation:
\(
\mathcal{R}_{\mathrm{SQL}} = \Big(\mathcal{G}_{\mathcal{LP}}, \{\mathcal{A}_i\}_{v_i \in V_{\mathcal{LP}}}\Big),
\)
which encodes both global semantic flow and the local syntactic details of each sub-SQL. 

\subsubsection{Context-Guided Property Embedding Fetching}
\label{sec:schema-guided-embedding}

After constructing the hierarchical intermediate representations (HIRs), we need to encode node attributes presented in text form from these HIRs. 
To capture semantically rich relationships between these heterogeneous attributes, we employ a pretrained LLM-based embedding model, augmented with contextual cues from the database schema and predicted SQL query.
This context-guided encoding forces the model to ground its representations in the specific database schema and SQL structure, aligning the semantic spaces of question, SQL syntax, and schema elements. 
This alignment is critical for bridging the inherent semantic gap between user queries and SQL logical structures, particularly for schema-specific components (e.g., column names, table relationships), where naive embeddings often fail to capture domain-specific nuances.

Specifically, we enhance each input sequence $x$ by prepending a \texttt{CONTEXT} prefix, forming an augmented input $x' = [\texttt{CONTEXT}; x]$ where $[\cdot~;~\cdot]$ denotes the concatenation function. 
% The \texttt{CONTEXT} is tailored to the input type:
% \ding{182} For natural language queries, \texttt{CONTEXT} is the database schema text (describing tables, columns, and their relationships).
% \ding{183} For node attributes from LPs/ASTs, \texttt{CONTEXT} is the complete predicted SQL query.
This augmented input $x'$ is fed into the embedding model $g(\cdot)$, which outputs a sequence of hidden states:
\begin{equation}
    H = g(x'), \quad \text{with } x' = [\texttt{CONTEXT}; x].
\end{equation}
We then extract task-specific embeddings from $H$ by taking the last hidden state corresponding to the $[\text{EOS}]$ token, yielding a condensed representation of the attribute's semantic features.
% based on the input type:
% \ding{182} For natural language queries, the embedding $h_{\mathrm{query}}$ is taken as the last hidden state corresponding to the $[\text{EOS}]$ token, capturing the holistic semantic representation of the query.
% \ding{183} For node attributes, the embedding is computed as the average of all hidden states in $H$, yielding a condensed representation of the attribute's semantic features.

% From $H$, we derive task-specific embeddings:
% - When $x$ is a natural language query, its embedding $h_{\mathrm{query}}$ is taken as the last hidden state of the query token sequence.
% - When $x$ is a node attribute, we collect the last hidden states of all attribute tokens to form an initial embedding set $H^{(0)} = \{h \mid h \text{ is the last hidden state of a node attribute token}\}$.

% Here, $h_{\mathrm{query}}$ captures the context-augmented semantic representation of the user's intent, while $H^{(0)}$ encodes the initial semantic features of structural components from LPs or ASTs.

% This context-guided encoding paradigm enforces the embedding model to condition its representations on the specific database schema and SQL structure, thereby aligning semantic spaces across natural language, SQL syntax, and schema elements. Critically, this alignment bridges the inherent semantic gap between user queries and SQL logical structures—particularly for tokens or attributes tightly bound to schema-specific components (e.g., column names, table relationships), where naive embeddings often fail to capture domain-specific nuances.

\subsubsection{Nested Message Passing Neural Network}
\label{sec:nested-MPNN}

% 基于Nested Message Passing Neural Network的SQL编码器：
After all node information has been encoded, we utilize a \textit{Nested Message Passing Neural Network} (NMPNN) to encode SQL queries by using the hierarchical structure of logical plans (LPs) and their associated abstract syntax trees (ASTs). 
This model consists of two consecutively applied message passing neural networks: a lower-level MPNN to aggregate node information within the AST in each LP node, and a higher-level MPNN to propagate and aggregate information across LP nodes.

\paragraph{\ding{182} Lower-level Message Passing within AST.}
The first stage of the NMPNN encodes the internal structure of each LP node by applying a message passing neural network over its corresponding AST. 
% Let $G_i=(V_i, E_i)$ denote the AST associated with logical plan node $i$, where $V_i$ and $E_i$ are the nodes and edges of the AST, respectively. 
% v_j^{\mathcal{A}_{i}} \in V_{\mathcal{A}_{i}}
The initial embedding $h_{v}^{(0)}\in H^{(0)}$ for each AST node $v \in V_{\mathcal{A}_{i}}$ is constructed from its syntactic type or token features. Message passing is performed for $T_\mathrm{AST}$ steps according to the standard MPNN update:
\begin{equation}
    h_{v}^{(t+1)} = \textsc{Update}_\mathrm{AST}\left(h_{v}^{(t)}, \textsc{Aggregrate}_\mathrm{AST}\left(\{h_{u}^{(t)}: u \in \mathcal{N}(v)\}\right)\right),
\end{equation}
where $\mathcal{N}(v)$ denotes the neighbors of $v$ in the AST. 
After $T_\mathrm{AST}$ steps, the node representations in the AST are aggregated via mean or sum pooling to produce an AST-level embedding $h_i = \textsc{Pooling}(\{h_{v}^{(T_\mathrm{AST})}\}_{v \in V_{\mathcal{A}_{i}}})$ for logical plan node $i$.

\paragraph{\ding{183} Higher-level Message Passing over Logical Plan.}
Once embeddings for all LP nodes are computed, each LP node $i \in V_{\mathcal{LP}}$ is now associated with an embedding $h_i$ derived from the corresponding AST. We apply an MPNN for $T_\mathrm{LP}$ iterations over the logical plan graph as follows:
\begin{equation}
    h_{i}^{(t+1)} = \textsc{Update}_\mathrm{LP}\left(h_{i}^{(t)}, \textsc{Aggregrate}_\mathrm{LP}\left(\{h_{j}^{(t)}: j \in \mathcal{N}_L(i)\}\right)\right),
\end{equation}
where $\mathcal{N}_L(i)$ denotes the set of neighboring nodes of $i$ in the logical plan. After $T_\mathrm{LP}$ steps, the final node embeddings are aggregated to form the representation $h_{\mathrm{SQL}}$ for the entire SQL query, which is $h_{\mathrm{SQL}} = \textsc{Pooling}(\{h_{i}^{(T_\mathrm{LP})}\}_{i \in V_{\mathcal{LP}}})$.

By hierarchically aggregating syntactic and semantic information from both the ASTs and the LP structure, the NMPNN is able to learn expressive representations for SQL queries, which enables effective modeling of the multi-level compositionality inherent in Text-to-SQL validation tasks.

\subsection{Data Augmentation for Text-to-SQL Validation}
\label{Data Augmentation for Text-to-SQL Validation}
% Section 4.2 针对训练数据不足的问题，介绍数据增广策略
% 章节总起段落先介绍ground truth数据\mathcal{D}从BIRD、Spider等数据集中获取，然后分别采用subsection 4.2.1 和 subsection 4.2.2 的策略扩充训练数据构建训练数据集
% subsection 4.2.1 介绍基于LLM的数据扩充：利用LLM根据ground truth数据中的question随机采样生成N条候选SQL，从中过滤掉存在语法错误的样本之后，与ground truth结果一致的为valid SQL，不一致的为invalid SQL
% subsection 4.2.2 介绍基于AST的数据扩充：从valid SQL中，利用AST变换规则进行语义扰动，构造大量负样本数据。

% A key challenge in training Text-to-SQL validation models lies in the limited scale of annotated datasets.

When training the SQL Validator, the availability of high-quality labeled data 
\(
    \mathcal{D}_{\mathrm{gold}} = \left\{ (q_i, s_i) \mid i \in \mathbb{N}^+, y_i = 1 \right\}
\)
is severely limited, and fine-grained annotations at the sub-SQL level are largely missing. 
Relying solely on the existing dataset is, therefore, insufficient to capture the diversity and complexity required for robust learning.  
To address this data scarcity issue, we introduce two complementary data augmentation strategies:  
\ding{182} \textbf{LLM-based data augmentation} (see Section~\ref{sec:llm_augmentation}); and  
\ding{183} \textbf{AST-driven sub-SQL augmentation strategy} (see Section~\ref{sec:ast_augmentation}).  
Together, these strategies enable us to construct a more diverse, challenging, and fine-grained training set, thereby enhancing both the robustness and generalization ability of the SQL Validator.  

\subsubsection{LLM-based Data Augmentation}
\label{sec:llm_augmentation}

LLMs exhibit remarkable generative capabilities, which can be harnessed for data augmentation. 
Specifically, for each question $q$ in the ground truth dataset $\mathcal{D}_{\mathrm{gold}}$, we prompt an LLM to generate $N$ candidate SQL queries $\{\tilde{s}_i\}_{i=1}^N$ by conditioning on question $q$.  Each generated SQL $\tilde{s}_i$ is then subjected to the following process:
\begin{itemize}[leftmargin=*,noitemsep,topsep=2pt]
    \item \textbf{\ding{182} Syntax Filtering}: We discard any syntactically invalid SQL queries that can be detected using an SQL compiler, and retain only those that can be parsed successfully for semantic validation.
    \item \textbf{\ding{183} Semantic Validation}: For the remaining queries, we compare the execution results of $\tilde{s}_i$ and the ground truth SQL query $s^{*} \in \mathcal{D}_{\mathrm{gold}}$. A candidate query $\tilde{s}_i$ is a \textit{valid SQL} if and only if
    $\mathrm{Exec}(\tilde{s}_i) = \mathrm{Exec}(s^{*}),$
    % \begin{equation}
    %     \mathrm{Exec}(\tilde{s}_i) = \mathrm{Exec}(s^{*}),
    % \end{equation}
    and is an \textit{invalid SQL} otherwise, where the function $\mathrm{Exec}(\cdot)$ denotes the output of executing a SQL query on the reference database.
\end{itemize}
After filtering and validation, we get LLM-based augmented training data $\mathcal{D}_{\mathrm{LLM}}$ both with diversified paraphrases of correct queries and with incorrect examples from semantically divergent outputs.

\subsubsection{AST-driven Sub-SQL Augmentation Strategy}
\label{sec:ast_augmentation}

To further enlarge the training corpus and introduce challenging semantic error samples, we leverage transformations on the AST representation of valid SQL queries. 
Concretely, for each valid SQL $s^{+}$ identified in Section~\ref{sec:llm_augmentation}, we systematically apply a set of AST-level transformation rules $\mathcal{T}$, designed to produce semantically perturbed variants. 
Each transformation $T \in \mathcal{T}$ maps $s^{+}$ to a mutated SQL $s^{-} = T(s^{+})$, such that the surface structure is altered while the semantic intent deviates from the original. 
\tealfont{By comparing the execution results between the augmented SQL query and ground truth SQL query, we can identify truly semantically incorrect examples.}
The details of the AST-level transformation rules $\mathcal{T}$ are provided in Section~\ref{sec:ast-rule}.  
Formally, the constructed semantic error sample set is:
\begin{equation}
    \mathcal{D}_{\mathrm{AST}} = \left\{(q, s^{-}) \:|\: s^{-} = T(s^{+}),\ T \in \mathcal{T},\ \mathrm{Exec}(s^{-}) \neq \mathrm{Exec}(s^{+}) \right\}.
\end{equation}

Beyond producing query-level semantic error samples, this augmentation also enables us to annotate \textit{sub-SQL fragments} at the LP level. 
Specifically, when a transformation perturbs a certain AST node, the corresponding sub-SQL in the LP can be marked as semantically incorrect, providing fine-grained supervision signals.  
This design enriches the training data with more informative error patterns, allowing the model to better capture subtle inconsistencies and to generalize more effectively in distinguishing between both global and local semantic errors.

% By integrating both LLM-based and AST-based augmentation strategies, we are able to construct a rich and balanced training dataset, significantly alleviating the data bottleneck and facilitating more effective learning of complex SQL generation tasks.

\subsection{Efficient Training of SQL Validation Model}
\label{Efficient Training of SQL Validation Model}
% SQL校验：将训练好的模型应用于SQL校验任务，通过计算用户查询和SQL语句在embedding层面的相似度分数，判断生成SQL是否符合用户查询意图，从而实现对SQL的语义校验

\paragraph{Fusion of Question and SQL Embeddings.}
To effectively integrate information from the natural language question and its corresponding SQL query, we use a multi-stage fusion strategy. 
\ding{182} First, given the question embedding $h_{\mathrm{question}} \in \mathbb{R}^d$ and the SQL embedding $h_{\mathrm{SQL}} \in \mathbb{R}^d$, we perform element-wise Hadamard multiplication~\citep{horn1990hadamard,horn2012matrix} to capture fine-grained interactions between the two modalities. The fused representation is given by $h_{\mathrm{hadamard}} = h_{\mathrm{question}} \odot h_{\mathrm{SQL}}$, where $\odot$ denotes the Hadamard product.
\ding{183} Then, we concatenate the original embeddings and the Hadamard fusion, forming $\mathbf{h} = [h_{\mathrm{question}}; h_{\mathrm{SQL}}; h_{\mathrm{hadamard}}] \in \mathbb{R}^{3d}$
, where $[\cdot\, ;\, \cdot\, ;\, \cdot]$ denotes vector concatenation. 
This concatenated vector aggregates both independent and interaction-based features of the input pair.
\ding{184} Finally, the combined vector $\mathbf{x}$ is fed into a three-layer multilayer perceptron (MLP) with $\mathrm{ReLU}$ activation~\citep{agarap2018deep} to produce the final output score $\hat{y}$. 
The process can be formally described as:
\begin{equation}
\hat{y} = \textsc{Mlp}([\, h_{\mathrm{question}} \,;\; h_{\mathrm{SQL}} \,;\; h_{\mathrm{question}} \odot h_{\mathrm{SQL}} \,]).
\end{equation}
This hierarchical fusion mechanism ensures that both shallow and deep interactions between question and SQL representations are captured, thereby facilitating effective SQL validation. 

\paragraph{Supervised Fine-Tuning for Text-to-SQL Validation.}
The Text-to-SQL validation task is formulated as a binary classification task, where the objective is to determine if a given SQL query is correct. 
Specifically, our model outputs a continuous score $\hat{y} \in [0,1]$ for each input instance. 
This score represents the predicted probability that the SQL is valid. During training, we employ the binary cross-entropy (BCE) loss to supervise the prediction, which is defined as follows:
\begin{equation}
    \mathcal{L}_{\mathrm{BCE}}(\hat{y}, y) = -[y \log(\hat{y}) + (1-y)\log(1-\hat{y})],
\end{equation}
where $y \in \{0, 1\}$ denotes the ground-truth label indicating whether the SQL query is invalid ($y=1$) or valid ($y=0$). 
Notably, after model training is complete, the embeddings aggregated at each node of the logical plan serve as valuable sources of semantic information.
Therefore, we can leverage the embeddings at logical plan nodes, together with the question embedding, to identify and localize erroneous sub-SQL segments produced by the LLM during the Text-to-SQL generation process. 
By providing targeted feedback based on the LP-level semantic signals, we can guide the LLM to recognize and correct specific errors in its intermediate outputs to iteratively refine its responses. 

\paragraph{Acceleration Strategies.}
To mitigate the time overhead associated with context-guided property embedding, there are two acceleration strategies being employed:
\ding{182} \textbf{KV Cache-based Schema Encoding.}
Since the schema information remains unchanged for a given database throughout the training process, we leverage the key-value (KV) cache mechanism~\citep{dao2022flashattention, dao2023flashattention2} to store intermediate representation results related to schema encoding. 
During repeated model invocations for the same database, previously computed KV cache entries of the schema can be directly reused, allowing the model to bypass redundant schema encoding computations. 
\ding{183} \textbf{Schema Information Compression.}
In the context of SQL semantic validation, detailed attributes for each schema column—such as data type, nullability, and primary key status—exert minimal influence on the semantic interpretation required for most downstream tasks. Therefore, to reduce the token footprint associated with schema representation, we retain only essential information: table names and corresponding column names. 
By filtering out non-essential metadata, we not only decrease the input length but also accelerate downstream processing, without notable loss in semantic adequacy for the intended validation tasks.

%% file: 5_Experiments.tex
\section{Experiments}

% 主实验：校验工作
    % backbone放主表 GAT、GCN、GIN
    
% 
% 
% 编码放不放schema信息
% 
% Case Study：证明C1
% 通用性实验：换一下Qwen/LLaMA/BERT 不同size
% 时间实验：LLM最长

In this section, we conduct comprehensive experiments on three Text-to-SQL datasets to evaluate the effectiveness of our method.
Further details and experiment results are provided in Section~\ref{sec: Further Experiment Details}.

% \begin{itemize}[leftmargin=*]
%     % 主要的模型效果分析
%     \item \textbf{RQ1} (\textbf{Section}~\ref{Main Results}, \textbf{Appendix}~\ref{appendix:Generalization Experiments}): 
%     Does \methodname surpass current SQL validation approachs under the same settings?
%     % 消融实验：消融：只有AST、只有LP、都没有 （C1）
%     \item \textbf{RQ2} (\textbf{Section} \ref{Ablation Study},\ref{EffiencyStudy}): 
%     Is \methodname effective in integrating both the global intent and the local structural details of SQL queries for better semantic information representation?
%     % 消融：不使用增强训练数据训练（C2）
%     \item \textbf{RQ3} (\textbf{Appendix}~\ref{Noise Poisoning Attack}, \ref{Case Study}): 
%     Can the AST-based negtive data augmentation strategy really help improve the robustness and generalization ability of model?
%     % 参数敏感性实验（C1） GNN层数（视野距离）
%     \item \textbf{RQ4} (\textbf{Appendix}~\ref{Sensitivity Analysis},~\ref{state_trend_study}): 
%     How sensitive and what trend is \M~to hyper-parameters $\sigma$ with different types $cppl$ and $uct$?
%     % case study
% \end{itemize}

\subsection{Experimental Setup}
\label{sec:exp setup}

\paragraph{Datasets.} 
We employ three datasets to evaluate the performance of Text-to-SQL validation, including two cross-domain Text-to-SQL datasets, \textit{BIRD}~\citep{li2024can} and \textit{Spider}~\citep{yu-etal-2018-spider}, and one medical domain dataset, \textit{EHRSQL}~\citep{lee2022ehrsql}, which is derived from queries about Electronic Health Records (EHR). 
% \olivefont{We further include the \textit{Ambrosia} \citep{saparina2024ambrosia} and \textit{Spider 2.0} \citep{leispider} datasets, using \textit{Ambrosia} as a related ambiguous-query benchmark and \textit{Spider 2.0} as an extended Spider variant featuring more complex schemas and SQLs.} 
\olivefont{To investigate more complex schemas and SQL queries, we also use the \textit{Spider 2.0} dataset~\citep{leispider}, which is much more difficult than \textit{BIRD} and \textit{Spider} datasets.}
More details of the \olivefont{four} datasets can be found in Section~\ref{sec:dataset}.

\paragraph{Methods and Baselines.}
In our experiments, we utilize two embedding models, namely \textit{Qwen3-Embedding-0.6B}~\citep{zhang2025qwen3} and \textit{embeddinggemma-300m}~\citep{gemmateam2025gemma3technicalreport}, for embedding-based methods. 
For LLM-based methods, we use the corresponding LLMS \textit{Qwen3-0.6B}~\citep{yang2025qwen3} and \textit{gemma-3-270m-it}~\citep{gemmateam2025gemma3technicalreport}.
To rigorously evaluate the effectiveness of our proposed approach, we compare \methodname against two categories of baselines: 
\ding{182} Baselines that treat SQL queries as plain text include Prompt, Chain-of-Thought (CoT)~\citep{wei2022chain}, ConfScore~\citep{somov2025confidence}, and COVE~\citep{dhuliawala2024chain}. 
\ding{183} To represent graph-based approaches which exploit the structural information of SQL queries, we select TED~\citep{chen2023error}, which leverages abstract syntax trees (ASTs) to model SQL query structure.
Implementation details for all baseline methods are provided in Section~\ref{sec:baselines}.

\paragraph{Experimental Settings.}
% %%% EXP SETTING
% 在训练阶段,我们合并了BIRD和Spider数据集中训练集的数据,形成基础的训练数据集。在此数据集的基础上利用Section 4.3提到的数据增强方法先后使用LLM-based的校验数据生成和AST-based的负样本数据增强方法扩充数据,得到数据增强后的训练集。而在测试阶段,我们分别测试模型在in-domain的数据集BIRD和Spider上的表现,以及out-domain的数据集EHRSQL上的表现。测试时的数据集仅包含LLM-based的校验数据生成的数据增强,不含AST-based的负样本扩增。评估模型性能时,我们选取了二分类评估中与阈值选择无关的两个指标AUPRC和AUROC作为评价指标。

During training, we combine the training splits of the \textit{BIRD} and \textit{Spider} datasets to construct a unified base training dataset, and apply the data augmentation techniques described in Section~\ref{Data Augmentation for Text-to-SQL Validation} to the dataset. 
Specifically, we first employ LLM-based data generation, followed by AST-driven sub-SQL augmentation, to expand the training data. 
For evaluation, we assess model performance on both in-domain datasets on \textit{BIRD} and \textit{Spider} and out-of-domain dataset \textit{EHRSQL} \olivefont{and \textit{Spider 2.0}}. 
% The in-domain performance is measured on the BIRD and Spider test sets, while out-of-domain generalization is evaluated using the EHRSQL dataset. 
Notably, during testing, we use only the LLM-based data augmentation for each dataset to match the validation scenario in the real world, excluding samples derived from AST-based negative data augmentation to ensure consistency in evaluation.
Detailed setting and evaluation of the experiments in Section~\ref{sec:exp-config}.

\paragraph{Evaluation Metrics.}
To measure semantic validation performance, we follow existing researches~\citep{chen-etal-2023-error} and adopt two threshold-independent metrics commonly used in binary classification: the \textit{Area Under the Precision-Recall Curve} (AUPRC) and the \textit{Area Under the Receiver Operating Characteristic Curve} (AUROC). 
These metrics provide a comprehensive assessment of model effectiveness regardless of the specific threshold chosen.

\begin{table*}[t]
% \color{olive}
\centering
\caption{Performance comparison (\%) on BIRD, Spider, EHRSQL and \olivefont{Spider 2.0} datasets. Results for methods with backbones of Qwen3-0.6B and Gemma-3-0.3B are presented in separate blocks for clarity. \textbf{Bold} indicates best results, and \underline{underline} indicates second best.}
\label{tab:spider_bird_ehrsql}
\fontsize{9pt}{10pt}\selectfont
\setlength{\tabcolsep}{4pt}
\renewcommand{\arraystretch}{1.25}
\resizebox{\linewidth}{!}{
% \begin{tabular}{l|cc|cc|cc|cc}
\begin{tabular}{l|cc|cc|cc|cc}
\toprule
\textbf{Method} 
& \multicolumn{4}{c|}{\textbf{In-Domain}} & \multicolumn{4}{c}{\textbf{Out-of-Domain}} \\ \cmidrule(lr){2-5} \cmidrule(lr){6-9}

& \multicolumn{2}{c|}{\textbf{BIRD}} 
& \multicolumn{2}{c|}{\textbf{Spider}} 
& \multicolumn{2}{c}{\textbf{EHRSQL}}
& \multicolumn{2}{c}{\textbf{\olivefont{Spider 2.0}}}
 \\
\cmidrule(lr){2-3} \cmidrule(lr){4-5} \cmidrule(lr){6-7}\cmidrule(lr){8-9}
& \textbf{AUPRC} & \textbf{AUROC}
& \textbf{AUPRC} & \textbf{AUROC}
& \textbf{AUPRC} & \textbf{AUROC} 
& \textbf{AUPRC} & \textbf{AUROC}
\\
\midrule

% ------------------- QWEN -------------------
\multicolumn{9}{c}{\textit{Qwen3-0.6B}} \\
\midrule
Prompt               & 60.85 & 57.25 & 40.86 & 58.94 & 84.72 & 58.86 &88.78 & 51.80\\
CoT                  & 61.01 & 55.31 & 39.58 & 50.98 & 84.13 & 59.61 & 89.15 & 48.32
 \\
ConfScore            & 56.59 & 51.43 & 34.13 & 50.65 & 80.66 & 50.39 & 90.15 & 44.77 \\
COVE                 & 57.67 & 54.40 & 38.09 & 54.78 & 85.53 & 56.54 & 90.75 & 61.86\\
TED                  & 56.28 & 52.37 & 33.14 & 46.63 & 76.07 & 41.76 & \underline{91.44} & 58.06\\
\methodname          & \textbf{67.39} & \textbf{61.51} & \textbf{51.92} & \textbf{67.32} & \textbf{89.07} & \textbf{69.53}& \textbf{92.59} & \textbf{64.32}\\
\quad - w/o AST \& LP & 62.66 & 59.23 & \underline{49.28} & 61.15 & 88.23 & 63.24  & 90.83 & 59.25 \\
\quad - w/o AST       & \underline{64.29} & 55.49 & 46.81 & \underline{62.73} & 87.34 & \underline{69.20} & {91.28} & {59.13} \\
\quad - w/o NDA       & 62.70 & \underline{59.25} & 48.91 & 59.97 & \underline{88.62} & 68.86 & 90.73 & \underline{59.36} \\
\midrule

% % ------------------- GEMMA -------------------
\multicolumn{9}{c}{\textit{Gemma-3-0.3B}} \\
\midrule
Prompt               & 54.13 & 48.47 & 34.26 & 49.75 & 78.86 & 44.79 & 88.30 & 44.58  \\
CoT                  & 54.72 & 49.84 & 34.28 & 49.96 & 81.73 & 54.11  & 89.22 & 50.00 \\
ConfScore            & 54.79 & 50.53 & 35.77 & 50.18 & 78.48 & 46.14& 90.79 & 52.34 \\
COVE                 & 55.02 & 50.46 & 34.25 & 49.78 & 81.58 & 51.79 & 88.51 & 45.86 \\
TED                  & 60.36 & 57.65 & 43.90 & 62.84 & 82.35 & 50.76 & 89.14 & 49.56 \\
\methodname          & \textbf{67.10} & \textbf{63.48} & \textbf{51.11} & \textbf{69.26} & \textbf{85.48} & \textbf{65.92}& \textbf{91.55} & \textbf{59.85} \\

\quad - w/o AST \& LP & \underline{66.71} & \underline{61.38} & \underline{50.39} & \underline{67.82} & 85.10 & 62.08 & 90.13  & \underline{54.49} \\
\quad - w/o AST       & 64.70 & 60.44 & 50.03 & 67.29 & \underline{85.22} & \underline{65.02} & \underline{90.90} & 53.28 \\
\quad - w/o NDA       & 64.39 & 60.56 & 49.73 & 66.19 & 84.14 & 60.77& 90.78 & 52.56  \\
\bottomrule
\end{tabular}
}
\end{table*}

\subsection{Text-to-SQL Validation Results}
\label{Main Results}

As discussed, comprehensive experiments are conducted and the evaluation results on the three datasets are shown in Table~\ref{tab:spider_bird_ehrsql}. 
The results lead to the following observations:

\paragraph{Outperforming SOTA Methods.}

% 介绍现象
Our proposed \methodname outperforms all baseline models across three Text-to-SQL validation datasets, confirming its superior ability to detect semantic discrepancies between questions and generated SQL queries. 
When evaluated with the in-domain datasets \textit{BIRD} and \textit{Spider}, \methodname achieves the highest AUPRC and AUROC scores among all methods, with average improvements of \textbf{16.28\%} (AUPRC) and \textbf{10.50\%} (AUROC) over previous state-of-the-art models. 
% 分析原因
One of the key reasons underlying this performance gain is that our method can effectively bridge global intent with local details. 
By capturing both the overarching semantic intent and the fine-grained local details within the SQL query, \methodname is able to perform more comprehensive and accurate cross-validation between the two modalities. 
% Such a dual-perspective approach enables our model not only to identify broad context mismatches but also to detect local inconsistencies and subtle errors that conventional methods often overlook. Consequently, \methodname delivers substantially improved performance in SQL query validation tasks.

\paragraph{Strong Semantic Validation Performance on Unseen Datasets.} 

As shown in Table~\ref{tab:spider_bird_ehrsql}, \methodname demonstrates strong semantic validation performance even on the clinical domain dataset \textit{EHRSQL} \olivefont{and complex dataset \textit{Spider 2.0}}, whose training sets did not appear during the training process of \methodname. 
\methodname reaches an average of \textbf{3.97\%} improvement in AUPRC and \textbf{19.34\%} in AUROC on the out-of-domain dataset \textit{EHRSQL}, 
highlighting its generalization ability without domain-specific fine-tuning. 
\olivefont{And on \textit{Spider 2.0} dataset, \methodname achieves an improvement of \textbf{1.05\%} in AUPRC and \textbf{9.16\%} in AUROC, demonstrating that it maintains strong performance even in scenarios with more complex schemas and SQL queries. This further confirms the robustness and generalization capability of \methodname across different domains and task complexities}
, making it a promising solution for practical deployment in diverse Text-to-SQL applications.
% This robustness on previously unseen medical data provides compelling evidence that our method can effectively bridge domain gaps and maintain high validation accuracy in real-world scenarios, where domain shift is commonly encountered. Overall, these findings suggest that \methodname is capable of delivering reliable validation performance on out-of-domain tasks

\subsection{Text-to-SQL Error Correction Aided by \methodname}

We conduct Text-to-SQL error correction experiments on the DEV set of the \textit{BIRD} dataset to evaluate the effectiveness of \methodname on improving the end-to-end performance of a Text-to-SQL pipeline. 
Our experiment compares end-to-end Text-to-SQL without self-correction, Prompt-based error detection method, and \methodname-based error detection method with and without LP-level feedback signals for error correction.
The experiments are evaluated using two backbone models, \textit{gpt-4o} and \textit{qwen-plus}, and the accuracy across three difficulty levels (simple, moderate, challenging) and the total accuracy are reported. 

\begin{table}[htbp]
\centering
\caption{Evaluation results with both \textit{gpt-4o} and \textit{qwen-plus} backbones for different Text-to-SQL error correction methods on the DEV set of \textit{BIRD} dataset under each difficult levels. \textbf{Bold} indicates best results, and \underline{underline} indicates runner-up results.}
\label{tab:verification_methods}
\renewcommand{\arraystretch}{1.25}
\setlength{\tabcolsep}{8pt}
\resizebox{\linewidth}{!}{
\begin{tabular}{lcccc}
\hline
\textbf{\quad \textbf{Method}} & \textbf{Simple} & \textbf{Moderate} & \textbf{Challenging} & \textbf{Total} \\
\hline
\rowcolor[HTML]{F2F2F2}
\multicolumn{5}{c}{\textit{gpt-4o}} \\

\quad W/o self-correction                      & \textbf{60.22} & 40.09 & 31.72 & 51.43 \\
\quad Prompt-base verifier                            & 57.51 & 38.36 & 29.66 & 49.09 \\
\quad \methodname-base verifier w/o LP-level feedback & \underline{59.24} & \underline{42.03} & \textbf{40.69} & \underline{52.28} \\
\quad \methodname-base verifier with LP-level feedback  & \textbf{60.22} & \textbf{43.97} & \underline{37.93} & \textbf{53.19} \\

\rowcolor[HTML]{F2F2F2}
\multicolumn{5}{c}{\textit{qwen-plus}} \\

\quad W/o self-correction                      & \underline{66.16} & 50.65 & 49.66 & 59.91 \\
\quad Prompt-base verifier                            & 64.00 & 48.28 & 49.66 & 57.89 \\
\quad \methodname-base verifier w/o LP-level feedback & 65.84 & \underline{53.02} & \textbf{51.72} & \underline{60.63} \\
\quad \methodname-base verifier with LP-level feedback  & \textbf{66.81} & \textbf{54.31} & \underline{51.34} & \textbf{61.47} \\
\hline
\end{tabular}
}   % resize
\end{table}

As shown in Table~\ref{tab:verification_methods}, applying \methodname-based error detection improves the end-to-end performance for Text-to-SQL tasks. 
For both non-thinking LLM backbone \textit{gpt-4o} and reasoning LLM backbone \textit{qwen-plus}, \methodname with LP-level feedback achieves the highest total accuracy, outperforming other baseline methods. 
Specially, \methodname-based verifier can significantly improve the performance of LLM for challenge tasks on BIRD dataset, gaining an improvement of \textbf{8.97\%} with no self-correction baselines. 
This improvement is mainly due to the fact that \methodname can effectively detect potential errors in the generated SQL, prompting the non-thinking model to reflect on the previously generated SQL and thereby correct the earlier errors.
Notably, introducing LP-level feedback further boosts performance. These results demonstrate that accurate error detection and LP-level guidance feedback are effective in enhancing the end-to-end performance of Text-to-SQL pipeline.
We also provide some case studies about Text-to-SQL error correction aided by \methodname in Appendix~\ref{sec:Case-Study}.

\subsection{Ablation Study}
\label{Ablation Study}

To validate the effectiveness of each component in \methodname, we conduct three ablation studies within our method. 
\ding{182} \methodname \textit{w/o NDA} model is trained on the dataset without AST-based negative data augmentation, while \ding{183} \methodname \textit{w/o AST} is the model that encodes the logical plan expression directly as text, without expanding it, and \ding{184} \methodname \textit{w/o AST \& LP} is the model just takes the entire SQL query as input.

The results of our ablation study, as shown in Table~\ref{tab:spider_bird_ehrsql}, demonstrate the effectiveness of each component in \methodname.
Removing NDA leads to a consistent decrease across all evaluation metrics, indicating that AST-based negative data augmentation helps the model better distinguish subtle semantic errors.
When the AST is excluded, the model mainly captures global intent from LP but loses the ability to understand important structural and syntactic details.
When both AST and LP are removed and the model is trained solely on plain text SQL queries, it becomes difficult for the model to capture either global intent or local grammatical and semantic details, resulting in performance decay.
These results verify the effectiveness of our approach in addressing challenges \textit{\textbf{\#C1}} and \textit{\textbf{\#C2}}.

%% file: 6_Conclusion.tex
\section{Conclusions and Future Works}
In this paper, we propose \methodname: a hierarchical representation framework that uses Logical Plans to capture high-level computational logic and data flow (global intent) \& Abstract Syntax Trees to model fine-grained schema details, and employs a Nested Message Passing Neural Network to aggregate schema-guided embeddings across relational AST/LP nodes. 
Moreover, an Adaptive sub-SQL Augmentation Strategy is introduced to generate large-scale syntactically valid but semantically incorrect negative samples via AST perturbations, thereby mitigating fine-grained data scarcity. 
Experiments on 
% \textit{BIRD}, \textit{Spider}, \textit{EHRSQL} and \olivefont{\textit{Spider 2.0}} (in-domain \& out-of-domain) 
both in-domain datasets and out-of-domain datasets
show that \methodname outperforms baselines in identifying semantic inconsistencies, which improves the reliability and interpretability of Text-to-SQL systems and reduces human verification costs. 
For future work, we will enhance the granularity of SQL validation by adding a fine-grained sub-SQL classification head that can not only detect semantic incorrectness in sub-SQL fragments but also classify specific error types (e.g., filter condition mismatches, aggregate function misselection, or join relation errors), achieving more precise and targeted semantic validation for Text-to-SQL queries.

%% file: 7_Appendix.tex
\newpage
\appendix

\startcontents[app]        % 从这里开始写入 buffer “app”

\section*{Appendix Table of Contents}      % 打印附录目录

\printcontents[app]{l}{1}{\setlength{\parindent}{0pt}}{}

\newpage

\input{8_7_notation}

\section{The Choice of Embedding Model Architecture}
\label{sec:choice-embedding}

In our work, we use embedding models with \textbf{decoder-only architecture} as the embedding backbone, rather than the encoder-only architecture that is more common in text encoding tasks. 
The choice of using decoder-only embedding models built upon large language models is because these models offer several distinct advantages. 

\ding{182} \textbf{Improved generalization ability}: 
With the development of large language models that are pretrained on vast and diverse corpora with next-token prediction objectives, these \textbf{decoder-only models can produce richer contextual representations}, enabling better transferability to downstream tasks and more robust performance across heterogeneous datasets. 
Embedding models like \textit{Qwen3-Embedding}~\citep{zhang2025qwen3} built upon \textit{Qwen3} series~\citep{yang2025qwen3} and \textit{EmbeddingGemma} built upon \textit{Gemma 3} series~\citep{gemmateam2025gemma3technicalreport} can have a better understanding of the unseen context in the training dataset.
\orangefont{In contrast, \textbf{open-source encoder-only models typically have far fewer parameters} than commonly used decoder-only models, which means they usually contain less knowledge.
For example, even the frequently used encoder-only architecture model \textit{GTE-large}~\citep{li2023towards} and \textit{BGE-large-en}~\citep{xiao2024c} have only around 0.3B parameters. 
As a result, these encoder-only models have relatively limited capacity to encode rich task-related knowledge. However, decoder-only models are available at significantly larger scales. Although we mainly use smaller models in our current experiments for computational efficiency, they can be readily replaced with larger variants when tackling more complex tasks.}

\ding{183} \textbf{Enhanced context length}: Modern decoder-only large language models are optimized to process significantly longer input sequences than their encoder-only counterparts.
For example, \textit{Qwen3-Embedding} series models support a maximum input sequence length of 32k tokens~\citep{zhang2025qwen3}, while \textit{BERT} series models are limited to a context length of only 512 tokens~\citep{devlin2019bert}.
The longer context allows models with decoder-only architecture to achieve richer semantic understanding over extended passages of text, \orangefont{which is crucial in realistic scenarios where a complex user query, database schema, and additional domain knowledge must be included in the context. 
Longer context windows make it easier to incorporate more detailed schema information, examples, and auxiliary hints, which is especially important in our \methodname framework}.

\orangefont{
To further verify these these, we conduct a experiment on \textit{BIRD}, \textit{Spider} and \textit{EHRSQL} datasets with both encoder-only architecture models and decoder-only architecture models.
In this experiment, we compare the performance of \methodname under difference embedding models, including ones with comparable parameter sizes (\textit{Qwen3-0.6B} and \textit{Gemma-3-0.3B}) as well as a larger variant (\textit{Qwen3-4B}), against strong encoder-only baselines (\textit{GTE-large} and \textit{BGE-large-en}) within the \methodname framework. The results are shown in Table~\ref{tab:embeddings}
}

% \begin{table}[ht]
% \color{orange}
% \centering
% \caption{
%     Comparison of the performance of \methodname under different embedding models.
% }
% \label{tab:embeddings}
% \resizebox{\linewidth}{!}{
% \renewcommand{\arraystretch}{1.2}
% \begin{tabular}{lcccccccc}
% \hline
% \multirow{2}{*}{\textbf{\shortstack{Embedding\\Model}}} & \multicolumn{2}{c}{\textbf{BIRD}} & \multicolumn{2}{c}{\textbf{Spider}} & \multicolumn{2}{c}{\textbf{EHRSQL}}& \multicolumn{2}{c}{\textbf{Spider2}} \\
% \cline{2-9}
%  & \textbf{AUPRC} & \textbf{AUROC} & \textbf{AUPRC} & \textbf{AUROC} & \textbf{AUPRC} & \textbf{AUROC} & \textbf{AUPRC} & \textbf{AUROC} \\
% \hline
% GTE-large         & 58.74 & 54.39 & 43.12 & 61.28 & 79.86 & 46.51&-&- \\
% BGE-large-en      & 62.01 & 57.78 & 42.87 & 61.87 & 81.41 & 49.34 &-&- \\
% Gemma-3-0.3b      & 67.10 & \underline{63.48} & 51.11 & \underline{69.26} & 85.48 & 65.92 &-&- \\
% Qwen3-0.6b        & \underline{67.39} & 61.51 & \underline{51.92} & 67.32 & \underline{89.07} & \underline{69.53}&92.59 & 64.32  \\
% Qwen3-4b          & \textbf{71.18} & \textbf{64.17} & \textbf{53.52} & \textbf{69.78} & \textbf{89.28} & \textbf{72.03}&-&-  \\
% \hline
% \end{tabular}
% }
% \end{table}
\begin{table}[ht]
% \color{orange}
\centering
\caption{
    Comparison of the performance of \methodname under different embedding models.
}
\label{tab:embeddings}
\resizebox{\linewidth}{!}{
\renewcommand{\arraystretch}{1.2}
\begin{tabular}{lcccccc}
\hline
\multirow{2}{*}{\textbf{\shortstack{Embedding\\Model}}} & \multicolumn{2}{c}{\textbf{BIRD}} & \multicolumn{2}{c}{\textbf{Spider}} & \multicolumn{2}{c}{\textbf{EHRSQL}}\\
\cline{2-7}
 & \textbf{AUPRC} & \textbf{AUROC} & \textbf{AUPRC} & \textbf{AUROC} & \textbf{AUPRC} & \textbf{AUROC}\\
\hline
GTE-large         & 58.74 & 54.39 & 43.12 & 61.28 & 79.86 & 46.51 \\
BGE-large-en      & 62.01 & 57.78 & 42.87 & 61.87 & 81.41 & 49.34 \\
Gemma-3-0.3b      & 67.10 & \underline{63.48} & 51.11 & \underline{69.26} & 85.48 & 65.92 \\
Qwen3-0.6b        & \underline{67.39} & 61.51 & \underline{51.92} & 67.32 & \underline{89.07} & \underline{69.53}  \\
Qwen3-4b          & \textbf{71.18} & \textbf{64.17} & \textbf{53.52} & \textbf{69.78} & \textbf{89.28} & \textbf{72.03}  \\
\hline
\end{tabular}
}
\end{table}

\orangefont{
Results from Table~\ref{tab:embeddings} show that decoder-only models with a parameter scale comparable to encoder-only models already achieve consistently better performance across all three datasets in \methodname. 
Moreover, scaling up the decoder-only model (e.g., from 0.6B to 4B) further improves performance, indicating a favorable scaling behavior for decoder-only embeddings in our setting.
}

\input{8_3_App_case}

\input{8_1_App_impl}

\input{8_2_App_exp}

\input{8_4_App_algo}

\input{8_8_sen}

\input{8_5_App_more}

\section{Data Ethics Statement}
To evaluate the efficacy of our work, we conducted experiments using \olivefont{5} datasets, including \textit{BIRD}, \textit{Spider}, \textit{EHRSQL}, \olivefont{\textit{Amrosia}, and \textit{Spider 2.0}}. All datasets are publicly available and used in accordance with their respective terms of use. No personally identifiable information was involved, and no human or animal subjects participated in this research.

\section{The Use of Large Language Models}

In this work, Large Language Models (LLMs) are utilized in polish writing and code development. 
In particular, LLMs helped refine the language and clarity of the paper, such as enhancing grammar and improving stylistic quality. 
LLMs were also employed to support the development of experimental code, including providing coding suggestions and troubleshooting assistance. 
All LLM-generated content was thoroughly reviewed and verified by the authors prior to inclusion. Research design, critical analyses, and all final decisions were carried out independently by the authors.

%% file: 8_7_notation.tex
\section{Notations Table}

\label{sec:notation}
The main notations in this paper are summarized in Table~\ref{tab:symbols}.

% \begin{table}[htb]
%     \centering
%       \caption{Notations Tables in \methodname}
%     \label{tab:symbols}
%     \begin{tabular}{c|l}        
%         \toprule
%         \rowcolor[gray]{0.94} Notation & Definition \\ 
%         \midrule
%         $q$ & natural language question \\
%         $s$ & SQL query \\
%         $\hat{y}$ & validation probability score \\
%         $y$ & validation label \\
%         $f$ & validation function \\
%         \hline
%         $h_{\mathrm{question}}$ & question embedding \\
%         $h_{\mathrm{SQL}}$ & SQL query embedding \\
%         $h_{\mathrm{hadamard}}$ & Hadamard fusion embedding  \\
%         \hline
%         $\mathcal{D}_{\mathrm{gold}}$ & ground truth Text-to-SQL dataset \\
%         $\mathcal{D}_{\mathrm{LLM}}$ & synthetic dataset with LLM-based data augmentation \\
%         $\mathcal{D}_{\mathrm{AST}}$ & synthetic dataset with AST-based negtive data augmentation \\
%         \bottomrule
%     \end{tabular}
%     % \vspace{-0.4cm}
% \end{table}

\begin{table}[htb]
    \centering
      \caption{Notations in \methodname}
    \label{tab:symbols}
    \begin{tabular}{c|l}        
        \toprule
        \rowcolor[gray]{0.94} Notation & Definition \\ 
        \midrule
        $q$ & natural language question \\
        $s$ & SQL query corresponding to question $q$ \\
        % $s^{*}$ & Ground truth SQL corresponding to question $q$ \\
        % $\tilde{s}$ & SQL query generated by the LLM conditioned on $q$ \\
        % $s^{+}$ & Valid SQL query (either ground truth $s^{*}$ or an LLM-generated semantically correct SQL) \\
        $\hat{y}$ & predicted validation probability score \\
        $y$ & validation label corresponding to question-SQL pair  \\
        $f$ & validation function \\
        \hline
        $\mathcal{LP}$ & logical plan \\
        $\mathcal{G}_{\mathcal{LP}}$ & directed acyclic graph of logical plan $\mathcal{LP} $\\
        ${V}_{\mathcal{LP}}$ & node sets of logical plan $\mathcal{G}_{\mathcal{LP}}$ \\
        ${E}_{\mathcal{LP}}$ & edge sets of logical plan $\mathcal{G}_{\mathcal{LP}}$ \\
        ${v}^{\mathcal{LP}}_{i}$ & node of logical plan $\mathcal{G}_{\mathcal{LP}}$ \\
        ${o}_{i}$ & relational operator of node ${v}^{\mathcal{L}}_{i}$\\
        ${a}_{i}$ & corresponding attributes of node ${v}^{\mathcal{L}}_{i}$\\
        $\mathcal{A}_i$ & Abstract Syntax Tree for node $i$ of $\mathcal{LP}$ \\
        $\mathcal{G}_{\mathcal{AST}}$ & directed acyclic graph of Abstract Synatx Tree $\mathcal{LP} $\\
        ${V}_{\mathcal{AST}}$ & node sets of AST $\mathcal{G}_{\mathcal{AST}}$  \\
        ${E}_{\mathcal{AST}}$ & edge sets of AST $\mathcal{G}_{\mathcal{AST}}$  \\  
        ${v}^{\mathcal{A}}_{j}$ & node j of AST $\mathcal{G}_{\mathcal{AST}}$ \\
        ${c}^{\mathcal{A}}_{j}$ & atom syntax of ${v}^{\mathcal{A}_i}_{j}$ \\
        $\mathcal{R}_{SQL}$ & hierarchical intermediate representation\\ %composite representation of hierarchical encoding method\\
        \hline
        % $x = (x_1, x_2, \dots)$ & Input sequence, where $x_i$ denotes the $i$-th element \\
        % $Context$ & schema text or a complete SQL query\\
        $g$ &  LLM-based embedding model \\
        $H$ & sequence of hidden states output by $g$ \\
        % $last$ & index set of the last hidden state for each node attribute \\
        % $H^0$ & initial representation of all AST node\\
        % $h_v^{(t)}$ & representation of AST node $v$ at message passing step $t$\\
        $h_v^{(t)}$ & representation of AST node $v \in V_{\mathcal{A}_i}$ at message passing step $t$ \\ % (with $h_v^{(0)}$ as its initial embedding) \\
        $\mathcal{N}(v)$ & neighbor set of node $v$ in AST $G_i$ \\
        $T_\mathrm{ast}$ & number of message passing steps in AST-level MPNN \\
        $h_i$ & aggregated embedding of LP node $i$ \\% , obtained by pooling over its AST nodes \\
        $h_i^{(t)}$ & representation of LP node $i$ at message passing step $t$ in logical-plan-level \\
        $\mathcal{N}_L(i)$ & neighbor set of LP node $i$ in the logical plan graph \\
        $T_\mathrm{LP}$ & number of message passing steps in logical-plan-level \\
        \hline
        $\mathcal{D}_{\mathrm{gold}}$ & ground truth Text-to-SQL dataset \\
        $\mathcal{D}_{\mathrm{LLM}}$ & synthetic dataset with LLM-based data augmentation \\
        $\mathcal{D}_{\mathrm{AST}}$ & synthetic dataset with AST-based negtive data augmentation \\
        % $i \in \mathbb{N}^+$ &Index of data pairs in the dataset \\
        $\mathbb{N}^+$ & index set of natural numbers (used to denote the size of $\mathcal{D}_{\mathrm{gold}}$) \\
        
        % $\mathrm{Exec}(\cdot)$ & execution function returning the output of a SQL query on the reference database \\
        % $\mathcal{T}$ & Set of AST-level transformation rules for perturbing valid SQL queries \\
        % $T \in \mathcal{T}$ &A single transformation rule applied to an AST \\
        $\mathcal{T}$ & set of AST-level transformation rules \\ % , where each $T \in \mathcal{T}$ denotes a single transformation\\
        % $s^{-} = T(s^{+})$ & Negative (semantically incorrect) SQL generated from applying transformation $T$ to $s^{+}$ \\
        % $(q, s^{-})$ & Question–SQL negative pair, where $\mathrm{Exec}(s^{-}) \neq \mathrm{Exec}(s^{+})$ \\
      \hline
        $h_{\mathrm{SQL}}$ & SQL query embedding \\
        $h_{\mathrm{question}}$ & question embedding \\
        $h_{\mathrm{hadamard}}$ & Hadamard fusion embedding  \\
      % $\odot$ & Hadamard (element-wise) product operator \\
      % $[,\cdot ,;, \cdot ,;, \cdot,]$ & Concatenation operator for vectors \\
      $\mathbf{h}$  & fused embedding vector \\% combining question, SQL, and interaction features \\
      % $\mathcal{L}_{\mathrm{BCE}}(\hat{y}, y)$ & Binary cross-entropy loss used to train the model \\
        \bottomrule
    \end{tabular}
    % \vspace{-0.4cm}
\end{table}

%% file: 8_3_App_case.tex
\section{Case Study on Text-to-SQL Error Correction Aided by \methodname}
\label{sec:Case-Study}
In this section, we present two detailed case studies selected from BIRD dataset to illustrate the problem-solving capability of \methodname when correcting erroneous SQL queries generated by a large language model (LLM). These examples demonstrate how \methodname effectively diagnoses the precise location and nature of the error in both the raw SQL and its logical plan representation, ultimately leading to the generation of the correct SQL.

As shown in Table~\ref{Case1}, the initial LLM output (\methodname Output) mistakenly adds WHERE T1. rtype = 'Reading'.  \methodname identifies this as an incorrect filter placement via the Logical Plan, where LogicalFilter(condition=[=(rtype, 'Reading')]) is applied after the JOIN.  The analysis highlights that the satscores table inherently contains only reading scores (as indicated by the schema not being fully utilized by the initial query logic or a subtle data characteristic), making the explicit WHERE clause redundant and potentially inefficient if multiple rtype values were involved.  The feedback correctly advises its removal, demonstrating that \methodname can detect and correct unnecessary/redundant operations, guiding the LLM to the concise and correct Ground Truth query.

\begin{table*}[htbp]
\centering
% \caption{\label{table:case_study_on_medical} Case Study on BIRD.}
\caption{Case study 1 on Text-to-SQL error correction process with the help of \methodname.}
\label{Case1}
\begin{tabular}{
  >{\centering\arraybackslash}m{2cm}
  >{\raggedright\arraybackslash}m{11.25cm}
}
\hline
\textbf{Question} &  For the school with the highest average score in Reading in the SAT test, what is its FRPM count for students aged 5-17 ? \\ \hline
\textbf{Ground Truth} & 
\begin{tcolorbox}[colback=blue!12!white,colframe=white,boxsep=0pt,left=0pt,right=0pt,top=0pt,bottom=0pt,after=\vspace{-1pt}]
SELECT T2.`FRPM Count (Ages 5-17)` \\
FROM satscores AS T1 \\
INNER JOIN frpm AS T2 \\
 ON T1.cds = T2.CDSCode \\
ORDER BY T1.AvgScrRead DESC \\
LIMIT 1;
\end{tcolorbox} \\ \hline
\textbf{\methodname Output} & 
\begin{tcolorbox}[colback=violet!12!white,colframe=white,boxsep=0pt,left=0pt,right=0pt,top=0pt,bottom=0pt,after=\vspace{-7pt}]
$<$LLM generate a error SQL$>$ \\
SELECT T2.`FRPM Count (Ages 5-17)` \\
FROM satscores AS T1 \\
INNER JOIN frpm AS T2 \\
\quad ON T1.cds = T2.CDSCode \\
\textcolor{red}{\textbf{WHERE T1.rtype = 'Reading'} }\\
ORDER BY T1.AvgScrRead DESC \\
LIMIT 1;
\end{tcolorbox}

\begin{tcolorbox}[colback=green!15!white,colframe=white,boxsep=0pt,left=0pt,right=0pt,top=0pt,bottom=0pt,after=\vspace{-7pt}]
$<$Anomalous node in Logical Plan$>$ \\
 LogicalProject(FRPM Count (Ages 5-17)=[`FRPM Count (Ages 5-17)`]) \\
\quad LogicalSort(sort0=[AvgScrRead], dir0=[DESC], fetch=[1]) \\
\quad\quad LogicalProject(FRPM Count (Ages 5-17)=[`FRPM Count (Ages 5-17)`], AvgScrRead=[AvgScrRead]) \\
\quad\quad\quad \textcolor{red}{\textbf{LogicalFilter(condition=[=(rtype, 'Reading')])} }\\
\quad\quad\quad\quad LogicalJoin(condition=[=(cds, CDSCode)], joinType=[inner]) \\
\quad\quad\quad\quad\quad JdbcTableScan(table=[[main, satscores]]) \\
\quad\quad\quad\quad\quad JdbcTableScan(table=[[main, frpm]]);
\end{tcolorbox}

\begin{tcolorbox}[colback=yellow!15!white,colframe=white,boxsep=0pt,left=0pt,right=0pt,top=0pt,bottom=0pt ,after=\vspace{-7pt}]
$<$LLM analyze based on error part$>$ \\
The mistake occurs because the filter `WHERE rtype = 'Reading'` is applied after the join in the predicted query and logical plan, rather than restricting the SAT scores table before aggregation or selection. In the logical plan, the `LogicalFilter(condition=[=(rtype, 'Reading')])` is placed after the join, which is unnecessary and inefficient because it forces the database to combine all rows from `satscores` and `frpm` before filtering only the Reading scores. This can lead to unnecessary data processing, larger intermediate result sets, and potentially incorrect selection if multiple `rtype` values exist for the same school. The core issue is that the filter should target the SAT scores table **before the join**, ensuring that only Reading scores are considered when identifying the school with the highest average, making the query both correct and efficient.
\end{tcolorbox}

\begin{tcolorbox}[colback=violet!12!white,colframe=white,boxsep=0pt,left=0pt,right=0pt,top=0pt,bottom=0pt,after=\vspace{-3pt}]
$<$LLM generate a correct SQL$>$ \\
SELECT T2.`FRPM Count (Ages 5-17)` \\
FROM satscores AS T1 \\
INNER JOIN frpm AS T2 \\
\quad ON T1.cds = T2.CDSCode \\
ORDER BY T1.AvgScrRead DESC \\
LIMIT 1;
\end{tcolorbox}
\\ \hline
\end{tabular}
\end{table*}

As shown in Table~\ref{Case2}, the LLM-generated SQL incorrectly uses WHERE schools.City = 'Fres', which the Logical Plan representation clearly shows as LogicalFilter(condition=[=(City, 'Fres')]). \methodname's analysis pinpoints a critical semantic error: the query filters by a non-existent City name ('Fres') instead of the required County Name ('Fresno'). This deep understanding of the question's intent versus the query's implementation allows \methodname to generate feedback that corrects both the column name (City to County Name) and the predicate value ('Fres' to 'Fresno'). This detailed, contextual error analysis enables the LLM to generate the final, logically correct SQL query that accurately addresses the user's intent.

\violetfont{As shown in Table~\ref{Case3}, the LLM-generated SQL incorrectly uses an INNER JOIN between schools and satscores, as highlighted in the Logical Plan with LogicalJoin(condition=[=(CDSCode, cds)], joinType=[inner]). \methodname’s analysis identifies a critical semantic mistake: the query assumes satscores.AvgScrWrite already contains the average writing score, ignoring the question’s requirement to compute the average over schools filtered by opening and closing dates. Moreover, the inner join would exclude schools without scores, leading to incomplete results. By understanding the intended aggregation and the need to preserve all eligible schools, \methodname generates feedback that corrects the join type to LEFT JOIN and clarifies that an explicit AVG aggregation is required. This detailed reasoning allows the LLM to produce a corrected SQL query that accurately computes the average writing scores for the specified subset of schools, while retaining optional phone numbers.}

\begin{table*}[htbp]
\centering
% \caption{\label{table:case_study_on_medical} Case Study on BIRD.}
\caption{Case study 2 on Text-to-SQL error correction process with the help of \methodname.}
\label{Case2}
\begin{tabular}{
  >{\centering\arraybackslash}m{2cm}
  >{\raggedright\arraybackslash}m{11.25cm}
}
\hline
\textbf{Question} & How many schools in Fresno (directly funded) have number of test takers not more than 250? \\ \hline
\textbf{Ground Truth} & 
\begin{tcolorbox}[colback=blue!12!white,colframe=white,boxsep=0pt,left=0pt,right=0pt,top=0pt,bottom=0pt,after=\vspace{-1pt}]
SELECT COUNT(T1.CDSCode) \\
FROM frpm AS T1 \\
INNER JOIN satscores AS T2 \\
\quad ON T1.CDSCode = T2.cds \\
WHERE T1.`Charter Funding Type` = 'Directly funded' \\
\quad AND T1.`County Name` = 'Fresno' \\
\quad AND T2.NumTstTakr $<=$ 250;
\end{tcolorbox} \\ \hline
\textbf{\methodname Output} & 
\begin{tcolorbox}[colback=violet!12!white,colframe=white,boxsep=0pt,left=0pt,right=0pt,top=0pt,bottom=0pt,after=\vspace{-7pt}]
$<$LLM generate a error SQL$>$ \\
SELECT COUNT(*) \\
FROM schools \\
JOIN satscores \\
\quad ON schools.CDSCode = satscores.cds \\
\textcolor{red}{\textbf{WHERE schools.City = 'Fres'}} \\
\quad AND schools.FundingType = 'Directly Funded' \\
\quad AND satscores.NumTstTakr $<=$ 250;
\end{tcolorbox} 

\begin{tcolorbox}[colback=green!15!white,colframe=white,boxsep=0pt,left=0pt,right=0pt,top=0pt,bottom=0pt,after=\vspace{-7pt}]
$<$Anomalous node in Logical Plan$>$ \\
LogicalAggregate(group=[{}], EXPR\$0=[COUNT()]) \\
\quad \textcolor{red}{\textbf{LogicalFilter(condition=[AND(=(City, 'Fres')}}, \\
\quad\quad =(FundingType, 'Directly Funded'), \\
\quad\quad $<=$(NumTstTakr, 250))]) \\
\quad\quad LogicalJoin(condition=[=(CDSCode, cds)], joinType=[inner]) \\
\quad\quad\quad JdbcTableScan(table=[[main, schools]]) \\
\quad\quad\quad JdbcTableScan(table=[[main, satscores]]);
\end{tcolorbox}

\begin{tcolorbox}[colback=yellow!15!white,colframe=white,boxsep=0pt,left=0pt,right=0pt,top=0pt,bottom=0pt ,after=\vspace{-7pt}]
$<$LLM analyze based on error part$>$ \\
The mistake occurs because the query filters `City = 'Fres'`, whereas the user’s intent is to count schools in Fresno County, not a city named “Fres.” In the logical plan, this corresponds to the `LogicalFilter(condition=[=(City, 'Fres')]…)` applied before the join. This condition is incorrect because `City` is the wrong column; the correct filter should be `'County Name' = 'Fresno'`. As a result, the query may return zero rows or only schools in a non-existent city “Fres,” completely missing schools that are actually in Fresno County. The redundant or incorrect part is the `City = 'Fres'` predicate, which should be replaced with `'County Name' = 'Fresno'` to correctly reflect the intended data selection. Without fixing this, the COUNT(\*) will not represent the number of directly funded schools in Fresno County with $\le$ 250 test takers.

\end{tcolorbox}

\begin{tcolorbox}[colback=violet!12!white,colframe=white,boxsep=0pt,left=0pt,right=0pt,top=0pt,bottom=0pt,after=\vspace{-3pt}]
$<$LLM generate a correct SQL$>$ \\
SELECT COUNT(*) \\
FROM schools \\
JOIN satscores \\
\quad ON schools.CDSCode = satscores.cds \\
WHERE schools.`County Name` = 'Fresno' \\
\quad AND schools.FundingType = 'Directly Funded' \\
\quad AND satscores.NumTstTakr $<=$ 250;
\end{tcolorbox}
\\ \hline
\end{tabular}
\end{table*}

\begin{table*}[htbp]
\centering
% \color{violet}
%\caption{\label{table:case_study_on_medical} Case Study on BIRD.}
\caption{\violetfont{Case study 3 on Text-to-SQL error correction process with the help of \methodname.}}
\label{Case3}
\begin{tabular}{
  >{\centering\arraybackslash}m{2cm}
  >{\raggedright\arraybackslash}m{11.25cm}
}
\hline
\textbf{Question} & What is the average score in writing for the schools that were opened after 1991 or closed before 2000? List the school names along with the score. Also, list the communication number of the schools if there is any. Communication number refers to phone number.
 \\ \hline
\textbf{Ground Truth} & 
\begin{tcolorbox}[colback=blue!12!white,colframe=white,boxsep=0pt,left=0pt,right=0pt,top=0pt,bottom=0pt,after=\vspace{-1pt}]
SELECT s.School, s.Phone, satscores.AvgScrWrite \\
    FROM schools AS s \\
       LEFT JOIN satscores \\
        ON s.CDSCode = satscores.cds \\
    WHERE s.OpenDate $>$ '1991-12-31' \\
        OR s.ClosedDate $<$ '2000-01-01'
\end{tcolorbox} \\ \hline
\textbf{\methodname Output} & 
\begin{tcolorbox}[colback=violet!12!white,colframe=white,boxsep=0pt,left=0pt,right=0pt,top=0pt,bottom=0pt,after=\vspace{-7pt}]
$<$LLM generate a error SQL$>$ \\
SELECT s.School, s.Phone, satscores.AvgScrWrite \\
FROM schools s \\
\textcolor{red}{\textbf{JOIN satscores}} \\
ON s.CDSCode = satscores.cds \\
WHERE s.OpenDate $>$ '1991-12-31' \\
OR s.ClosedDate $<$ '2000-01-01'
\end{tcolorbox} 

\begin{tcolorbox}[colback=green!15!white,colframe=white,boxsep=0pt,left=0pt,right=0pt,top=0pt,bottom=0pt,after=\vspace{-7pt}]
$<$Anomalous node in Logical Plan$>$ \\
LogicalProject(SCHOOL=[School], PHONE=[Phone], AVGSCRWRITE=[AvgScrWrite]) \\
  LogicalFilter(condition=[OR($>$(OpenDate, '1991-12-31'), $<$(ClosedDate, '2000-01-01'))]) \\
   \textcolor{red}{\textbf{ LogicalJoin(condition=[=(CDSCode, cds)], joinType=[inner])}} \\
      JdbcTableScan(table=[[main, schools]]) \\
      JdbcTableScan(table=[[main, satscores]])
\end{tcolorbox}

\begin{tcolorbox}[colback=yellow!15!white,colframe=white,boxsep=0pt,left=0pt,right=0pt,top=0pt,bottom=0pt ,after=\vspace{-7pt}]
$<$LLM analyze based on error part$>$ \\
The mistake occurs because the predicted SQL and logical plan treat `satscores.AvgScrWrite` as if it already contains the “average writing score,” whereas the question actually asks for calculating the average writing score for the subset of schools that meet the date criteria. In the query, the `INNER JOIN` between `schools` and `satscores` is correct for matching CDS codes, but it does not account for schools that might lack scores—here a `LEFT JOIN` would ensure all eligible schools are included even if no scores exist. Moreover, the query lacks an `AVG(satscores.WriteScore)` aggregation over the filtered subset; directly selecting `AvgScrWrite` is incorrect and redundant. The implication is that the query may either exclude schools without scores (due to the inner join) or return precomputed averages that do not reflect the filtered date condition, producing inaccurate results. In short, the query should use a `LEFT JOIN` and an explicit `AVG` aggregation over the writing scores of schools that satisfy the opening/closing date condition.

\end{tcolorbox}

\begin{tcolorbox}[colback=violet!12!white,colframe=white,boxsep=0pt,left=0pt,right=0pt,top=0pt,bottom=0pt,after=\vspace{-3pt}]
$<$LLM generate a correct SQL$>$ \\
SELECT s.School, s.Phone, satscores.AvgScrWrite \\
    FROM schools s \\
LEFT JOIN satscores \\
        ON s.CDSCode = satscores.cds \\
    WHERE s.OpenDate $>$ '1991-12-31' \\
        OR s.ClosedDate $<$ '2000-01-01'
\end{tcolorbox}
\\ \hline
\end{tabular}
\end{table*}

%% file: 8_1_App_impl.tex
\section{Implementation Details}

\subsection{Baseline Implementation}
\label{sec:baselines}

\begin{itemize}[leftmargin=*,noitemsep,topsep=2pt]
    \item {Prompt.} We directly prompt a large language model (LLM) to determine whether a natural language question and an SQL query are semantically aligned. 
    \tealfont{Given a user's natural language question and the corresponding database schema, we request the LLM to check whether the generated SQL query is semantically correct based on a list of validation rules. We provide the validation prompt in Section~\ref{sec: prompts}} 
    \item {CoT.} We extend Prompt method by applying chain-of-thought prompting~\citep{wei2022chain}, allowing the model to reason step by step before deciding whether the text and SQL match. 
    \tealfont{Specifically, instead of requiring the LLM to provide a direct judgment, we instruct the model to explicitly enumerate the semantic intent expressed in the user's question, to analyze the components of the SQL query, and to compare their correspondence according to a set of validation criteria.}
    \item {ConfScore.} We build on the confidence-based evaluation method of \citet{somov2025confidence}, originally proposed for SQL generation. To adapt it to the SQL verification setting, we further draw on the prompting strategy of \citet{kim2023sure}, enabling the model to take both the text and candidate SQL as input and produce an explanation whose confidence is then used to assess correctness.
    \item {COVE.} Inspired by \citet{dhuliawala2024chain}, COVE reduces hallucinations through verification and correction. While the original pipeline involves generating an initial answer and then validating it, in our SQL verification setting we skip the answer generation step and directly apply the subsequent stages: (i) planning verification questions from the text–SQL pair, (ii) answering these questions independently, and (iii) aggregating the verification results to assess whether the SQL is correct.  
    \item {TED.} Text-to-SQL Error Detector (TED)~\citep{chen2023error} is a parser-agnostic model that jointly encodes natural language questions and SQL queries using CodeBERT~\citep{Feng2020CodeBERT}, while incorporating graph neural networks (GNN) to capture structural information. By modeling dependency trees of questions and abstract syntax trees (ASTs) of SQL queries, TED explicitly learns the semantic alignment between text and query structures, enabling more accurate detection of mismatches.  
\end{itemize}

\subsection{Details about AST-Level Transformation Rules}
\label{sec:ast-rule}

To construct a series of challenging negative samples from valid SQL queries, we define a set of AST-level transformation rules $\mathcal{T}$ to generate samples that are grammatically correct but semantically incorrect, which is introduced in Section~\ref{sec:ast_augmentation}. All transformation rules leveraged in our work are shown below.

\begin{itemize}[leftmargin=*]
    \item \textbf{Rule 1: Operator Inversion.} 
    Many SQL queries rely on logical and comparison operators to specify selection conditions. 
    To perturb the query's semantics while retaining grammatical validity, we invert operators at the AST level. 
    For example, greater-than operators ($>$) are replaced with less-than-or-equal-to ($\leq$), and logical connectors such as \texttt{AND} are substituted with \texttt{OR}. 
    This transformation ensures the resulting query is syntactically correct, but alters the retrieval logic to produce incorrect results.

    \item \textbf{Rule 2: Identifier Substitution.}
    In this rule, we randomly select field identifiers from the same database schema and replace the original field names in the SQL query's AST. 
    For instance, replacing column name `salary` with `age`, or replacing table name `department` with `location`. 
    The replacement fields are drawn from the candidate column names pool or table names pool to ensure grammatical correctness, but disrupt the intended query semantics.

    \item \textbf{Rule 3: Constant Replacement.}
    We modify constant values appearing in the SQL conditions by substituting them with other constants from the database. Suppose a condition checks for ``age = 30''; we may replace 30 with a different valid value such as 40. The new constants are selected to be plausible for the context but alter the outcome of the query.

    \item \textbf{Rule 4: Aggregation Function Mutation.}
    Aggregation functions define how rows are combined in SQL queries. We alter these functions by replacing one aggregation operator with another, e.g., substituting AVG with MAX, MIN, or COUNT. This transformation keeps the query structurally intact but changes its fundamental semantic intention.

\end{itemize}

These rules constitute a comprehensive transformation set designed for AST-based SQL augmentation. By applying $\mathcal{T}$ to source queries, we generate a variety of negative samples that challenge models on fine-grained semantic understanding without violating SQL syntax. The full instantiation of each rule is implemented over the SQL AST, allowing flexible and systematic query perturbation for downstream training and evaluation.

% 介绍数据集
\subsection{Construction and Splitting Strategy of Datasets}

As for the datasets, we follow the original train/test splits for all three datasets: \textit{BIRD}, \textit{Spider}, and \textit{EHRSQL}. 
During training, we combine the training splits of the \textit{BIRD} and \textit{Spider} datasets to construct a unified base training dataset. 
Building upon this merged dataset, we apply the data augmentation techniques described in Section~\ref{sec:llm_augmentation}. 
Specifically, we first employ LLM-based data validation generation, followed by AST-driven sub-SQL augmentation strategy that introduced in Section~\ref{sec:ast_augmentation}, to expand the training data. 
To balance the ratio of positive and negative samples in the training data more effectively, we imposed a restriction on the number of augmented samples during the AST-driven sub-SQL augmentation process, ensuring that the ratio of positive to negative samples in the final training set could reach 1:1.
The final augmented training set thus integrates both LLM-generated validation data and augmentated semantic error samples, and we split the final data into training and validation sets using a ratio of 80\% to 20\% .

\tealfont{
Notably, \textbf{all samples generated by the AST-driven sub-SQL augmentation strategy are truely semantically incorrect}. 
Concretely, our training set is constructed from SQL queries that have known ground-truth SQL counterparts, which can be executed to obtain ground-truth execution results. 
After applying AST perturbations, we execute each perturbed SQL query and compare its execution result with that of the corresponding ground-truth query. Only perturbed SQL queries whose execution results differ from the original are retained as semantically incorrect samples in the augmented dataset. 
This strict, result-based filtering, ensemble reverse reject-sampling, substantially reduces the likelihood of including false negatives and ensures that retained perturbations correspond to genuine semantic or logical errors.
}

\tealfont{
Regarding overfitting to synthetic mistakes, the AST-perturbed negative samples are always used in conjunction with the original (question, SQL) pairs and other negative examples in our augmented corpus. In practice, this means the model is exposed to a diverse mixture of real and systematically perturbed errors, rather than being trained solely on a narrow class of synthetic perturbations. This diversity, together with the execution-based filtering, helps mitigate overfitting and encourages the model to learn robust semantic distinctions rather than artifacts of a particular perturbation scheme.
}

In the evaluation process, we assess model performance on both in-domain and out-of-domain datasets. The in-domain performance is measured on the \textit{BIRD} and \textit{Spider} test sets, while out-of-domain generalization is evaluated using the \textit{EHRSQL} dataset. 
Notably, during testing, we use only the LLM-based validation data augmentation for each dataset, excluding samples derived from AST-driven sub-SQL augmentation to ensure consistency in evaluation.

% \begingroup
% \color{magenta}
\subsection{Threshold Selection for Real-World Applications}

The binary validator decision threshold in our Text-to-SQL pipeline should be chosen according to the requirements of the downstream applications. 
Improved AUROC/AUPRC implies a better Pareto frontier, enabling practitioners to select more favorable precision–recall trade-offs for different risk profiles. Concretely:
\begin{itemize}[leftmargin=*]
    \item In scenarios where \emph{false positives (FP)} are more harmful (e.g. over-correcting originally correct SQL and potentially regressing performance), one should choose a higher threshold that prioritizes precision, even at the cost of recall. In practice, this can be done by selecting the largest threshold at which precision stays above a desired target (e.g., $\geq 95\%$). 
    \item In scenarios where \emph{false negatives (FN)} are more costly (e.g. it is critical not to miss any erroneous SQL being executed), one may prefer a lower threshold that emphasizes high recall, while allowing precision to vary within an acceptable range.
\end{itemize}

In our experiments, we report results using the threshold that maximizes the F1 score on a validation set, providing a balanced operating point between precision and recall. This is a standard choice when there is no application-specific preference between FP and FN. We also illustrate that, under a reasonable threshold for error detection and targeted correction, integrating our validator can improve end-to-end Text-to-SQL accuracy.

% \endgroup

\subsection{Prompt Used in the Framework}
\label{sec: prompts} 

In this section, we provide a detailed introduction to the prompts used in our framework.

% LLM-based augmentation
\begin{tcolorbox}
[colback=lightgray!20,colframe=darkgray!80,title=LLM-based Data Augmentation Prompt]

You are a helpful data analysis expert. Given a user question and the corresponding database schema, please output a syntactically correct and efficient SQL query that fully answers the user's request.\\

The database schema is as follows:

\texttt{\{schema\}}\\

Write SQL for the following question:

\texttt{\{question\}}\\

Please output the SQL query directly. Do NOT output any other comments.

\end{tcolorbox}

% LLM Verification
\begin{tcolorbox}
[colback=lightgray!20,colframe=darkgray!80,title=LLM Verification Prompt]

\textbf{[Role]}

You are an expert specializing in SQL query verification, dedicated to the task of determining whether a given SQL can solve the user's problem, and providing high-quality analysis results based on specific rules. You possess strong knowledge of SQL syntax, understanding of database structures, and natural language processing abilities, allowing you to accurately judge the match between SQL queries and user questions.

\vspace{\baselineskip}

\textbf{[Workflow]}

\begin{enumerate}[leftmargin=*]
    \item \textbf{Input Acquisition} \\
      \textbf{Receive} three essential pieces of information from the user:
      \begin{itemize}
          \item \textbf{User Question (question)}
          \item \textbf{Database Structure (schema)}
          \item \textbf{SQL Query Statement}
      \end{itemize}
      Ensure successful parsing and prepare for detailed analysis.
      
    \item \textbf{Analysis / Processing Logic}
      \begin{enumerate}[label*=\arabic*.]
        \item \textbf{Question Analysis:} \\
            Analyze the semantic intent of the user question, identifying key \textbf{entities}, \textbf{attributes}, and \textbf{query conditions}.
            
        \item \textbf{Database Structure Analysis:} \\
            Understand the \textbf{table structure}, \textbf{field types}, and \textbf{relationships} within the schema. Determine which tables and fields are relevant to the user question.
            
        \item \textbf{SQL Syntax Check:} \\
            Verify the syntactic correctness of the SQL query, including:
            \begin{itemize}
                \item keyword usage 
                \item table and field references
                \item JOIN syntax, etc.
            \end{itemize}
            
        \item \textbf{Semantic Matching Analysis:}
            \begin{itemize}
                \item \textbf{SELECT} clause: Check if all fields required to answer the question are included.
                \item \textbf{FROM} \& \textbf{JOIN} clauses: Ensure all relevant tables are referenced.
                \item \textbf{WHERE} clause: Confirm inclusion of all necessary filtering conditions.
                \item \textbf{GROUP BY}, \textbf{HAVING}, \textbf{ORDER BY}, etc.: Verify compliance with the question’s requirements.
            \end{itemize}
            
        \item \textbf{Result Expectation Analysis:} \\
            Infer whether the result set after SQL execution can directly answer the user question, or if further processing is needed.
      \end{enumerate}

    \item \textbf{Result Output} \\
      Based on the analysis results, provide a clear judgment:
      \begin{center}
      \fcolorbox{black}{gray!15}{\textbf{approve}} \quad \textbf{or} \quad \fcolorbox{black}{gray!15}{\textbf{reject}}
      \end{center}
      \textit{Answer “approve” or “reject” only, without any additional explanation.}
\end{enumerate}

% \vspace{\baselineskip}

% \textbf{[Rules]}

% Rule 1: SQL Syntax Correctness and Database Structure Compatibility - The SQL query must be syntactically correct, and all referenced tables and fields must exist in the database schema. If the SQL query contains syntax errors or references to non-existent tables/fields, the judgment is "reject".

% Rule 2: Semantic Matching of SQL Query and User Question - The SQL query must be able to extract all the information required to answer the user question. If the SQL query misses key conditions or query targets mentioned in the question, the judgment is "reject".

% Rule 3: Completeness and Accuracy of SQL Query Results - The result of the SQL query must completely and accurately answer the user question, without redundant or missing information. If the result set returned by the SQL query contains redundant information or lacks necessary information, the judgment is "reject".

% Rule 4: Special Case Handling - For questions requiring multi-table joins, nested queries, aggregate functions, and other complex SQL features, you must ensure that the SQL query correctly implements these features. If the question requires these features but the SQL query does not implement them correctly, the judgment is "reject".

\vspace{\baselineskip}
\textbf{[Question]}

\texttt{\{question\}}

\vspace{\baselineskip}
\textbf{[Database Schema]}

\texttt{\{schema\}}

\vspace{\baselineskip}
\textbf{[SQL]}

\texttt{\{SQL\}}

% \textbf{Output Format}

% Please provide your verification result by answering "approve" or "reject" directly. Do NOT output any other words.

\end{tcolorbox}

% LLM-based analyze
\begin{tcolorbox}
[colback=lightgray!20,colframe=darkgray!80,title=LLM Analyze Based on Error Part Prompt]

You are an expert SQL analyst.   Carefully examine the given SQL query, its logical plan, and the described mistake.   Provide a detailed, step-by-step explanation in English.\\

Question:\\
\texttt{\{question\}}\\

Predicted SQL query:\\
\texttt{\{pred\_sql\}}\\

Predicted logical plan:\\
\texttt{\{pred\_plan\}}\\

Identified mistake of logical plan:\\
\texttt{\{pred\_mistake\}}\\

Instructions for analysis:\\
1. Explain clearly why the mistake occurs.\\
2. Identify which part of the query is incorrect, redundant, or unnecessary.\\
3. Describe the implications of this mistake on query execution or results.\\
4. Provide a concise, understandable explanation suitable for someone familiar with SQL and query planning.\\

Output:\\
A detailed error analysis, focusing only on the predicted SQL and its logical plan. Please generate a short paragraph.

\end{tcolorbox}

% % LLM correction prompt
% \begin{tcolorbox}
% [colback=lightgray!20,colframe=darkgray!80,title=Text-to-SQL Error Correction Prompt for LLM with the Help of \methodname]

% \end{tcolorbox}

%% file: 8_2_App_exp.tex
\section{Experiments Details}
\label{sec: Further Experiment Details}

\subsection{Statistical Information of Datasets}
\label{sec:dataset}

\begin{table}[ht]
\centering
\caption{Overview of datasets used in our experiments.}
\label{tab:dataset_overview}
\fontsize{9pt}{9pt}\selectfont
\setlength{\tabcolsep}{6pt}
\renewcommand{\arraystretch}{1.1}
\begin{tabular}{l|c|c|c|c|c|c}
\toprule
\rowcolor{gray!10}
\textbf{Dataset} & \textbf{Domain} & \textbf{\#DB} & \textbf{\#Table/DB} & \textbf{Train} & \textbf{Dev} & \textbf{Test} \\
\midrule
BIRD~\citep{li2024can}          & 37 & 95  & 7.3 & 9,428    & 1,534   & 1,789 \\
Spider~\citep{yu-etal-2018-spider}      & 138 & 200 & 5.1 & 7,862 & 1,831 & 2,147 \\
EHRSQL~\citep{lee2024overview} & 1 & 1 & 17 & 5,124 & 1,163 & 1,167 \\
Spider 2.0~\citep{saparina2024ambrosia} & / & 213 & 7.5 & / & / & 632  \\
Ambrosia~\citep{saparina2024ambrosia} & 16 & 846 & 5.0 & / & / & 1,277 \\
\bottomrule
\end{tabular}
\end{table}

In our work, five different datasets are taken to evaluate Text-to-SQL validation performance: two general-purpose datasets, \textit{BIRD}~\citep{li2024can} and \textit{Spider}~\citep{yu-etal-2018-spider}, one practical medical dataset \textit{EHRSQL}~\citep{lee2022ehrsql,lee2024overview}, \olivefont{one challenge dataset \textit{Spider 2.0}~\citep{leispider} and one ambiguous dataset \textit{Ambrosia}~\citep{saparina2024ambrosia}}.

\begin{itemize}[leftmargin=*]
    \item \textbf{BIRD}~\citep{li2024can} is a large-scale cross-domain benchmark designed for real-world Text-to-SQL tasks. 
    It contains 12751 questions and their corresponding SQL queries, grounded in 95 databases across 37 domains.
    The dataset is divided into a training set with 9,428 instances, a development set with 1,534 instances, and a test set with 1,789 instances.
    BIRD's unique focus on large-scale databases and external knowledge makes it a challenging benchmark for real-world Text-to-SQL tasks.
    
    \item \textbf{Spider}~\citep{yu-etal-2018-spider} is also a cross-domain Text-to-SQL dataset. 
    It contains 10,181 questions and 5,693 unique complex SQL queries across 200 databases across 138 domains.The dataset is evaluated under two settings: (i) example split, where 7,862/1,831/2,147 questions are randomly assigned to train/dev/test, allowing questions from the same database to appear across splits; and (ii) database split, where 206 databases are divided into 130/36/40 for train/dev/test, ensuring all questions from a database remain within the same split. Spider poses greater challenges than prior text-to-SQL datasets due to its large number of complex queries (e.g., joins, nesting, grouping, ordering) and its requirement for cross-domain generalization to unseen database schema.
    \item \textbf{EHRSQL}~\citep{lee2022ehrsql,lee2024overview} is a Text-to-SQL dataset focused on electronic health records (EHR), grounded in real-world database, MIMIC-IV~\citep{johnson2023mimic}. Specifically, we use the EHRSQL 2024 dataset in our experiments.
    The dataset is split into a training set with 5124 questions, a development set with 1163 questions, and a test set with 1167 questions.   
    A unique feature is the inclusion of unanswerable questions, which requires models to leverage external knowledge to determine question validity.
    \item \olivefont{
    \textbf{Spider 2.0}~\citep{leispider} is a large-scale, cross-domain Text-to-SQL benchmark that significantly scales up in database size, query complexity, and SQL dialect diversity compared to its predecessor. It comprises 632 test examples across 213 databases, with an average of 743.5 columns per database and 148.3 tokens per SQL query, reflecting a high degree of schema complexity and query length. Each SQL query involves an average of 7.1 functions, indicating extensive use of advanced SQL constructs. The databases span a wide range of real-world domains and include diverse data types like JSON, STRUCT, GEOGRAPHY, and TIMESTAMP across multiple SQL dialects, which include BigQuery, Snowflake and SQLite. Notably, Spider 2.0 databases are TB-scale, with an average of 5.27 billion rows, presenting unprecedented challenges in scale, schema understanding, and cross-domain generalization for Text-to-SQL systems.
    In our experiments, we evaluate models under the Spider 2.0-Lite subset using SQLite as the database engine.
    }
    \item \olivefont{\textbf{Ambrosia}~\citep{saparina2024ambrosia} is a cross-domain benchmark dedicated to parsing \textbf{ambiguous} natural-language questions into SQL. It contains 1,277 ambiguous questions, 2,965 corresponding SQL interpretations, and 846 multi-table databases spanning 16 realistic domains (e.g., Banking, Entertainment, Healthcare). Each ambiguous question is paired with all valid, human-verified SQL queries that arise from scope, attachment, or vagueness ambiguities, yielding 2–3 gold queries per question. The dataset is split 90 \% for zero shot evaluation and 10 \% for few-shot demonstration; questions are grouped by ambiguity type rather than by database to preserve linguistic variety. Compared with prior text-to-SQL corpora, Ambrosia introduces the first large-scale testbed where ambiguity persists even when the full schema and content are known, requiring models to explicitly recognize and generate multiple correct interpretations instead of a single canonical query.}
\end{itemize}

\magentafont{
In our experiments, the class distribution of training set is balanced with 1:1 ration because we applied negative-sample augmentation strategy for the training set.
For the evaluation set, we analyzed the class distribution in each dataset used in our main experiments. 
}

% \begin{table}[ht]
% \centering
% \caption{Class distribution of test sets across datasets}
% \label{tab:class_distribution}
% \fontsize{9pt}{9pt}\selectfont
% \setlength{\tabcolsep}{8pt}
% \renewcommand{\arraystretch}{1.1}
% \begin{tabular}{l|c|c|c}
% \toprule
% \rowcolor{gray!10}
% \textbf{Dataset} & \textbf{\#Correct} & \textbf{\#Incorrect} & \textbf{Positive Ratio (\%)} \\
% \midrule
% BIRD~\citep{li2024can}    & 812  & 983  & 54.76 \\
% Spider~\citep{yu-etal-2018-spider} & 1019 & 532  & 34.30 \\
% EHRSQL~\citep{lee2024overview}     & 141  & 579  & 80.41 \\
% \bottomrule
% \end{tabular}
% \end{table}

% \begin{wrapfigure}{r}{0.5\textwidth}  
%     \vspace{-0.3cm}  % 调整上方间距
%     \centering
%     \includegraphics[width=0.46\textwidth]{figures/dataset.pdf}
%     \caption{Visualization of class distribution of test sets across datasets.}
%     \label{fig:testset_distribution}
%     \vspace{-0.3cm}  % 调整下方间距
% \end{wrapfigure}

\begin{figure}[ht]
% \color{magenta}
  \centering
  \begin{minipage}[c]{0.5\textwidth}
    \centering
    \captionof{table}{Class distribution of test sets across datasets}
    \label{tab:class_distribution}
    \fontsize{7pt}{7pt}\selectfont
    \setlength{\tabcolsep}{8pt}
    \renewcommand{\arraystretch}{1.1}
    \begin{tabular}{l|c|c|c}
      \toprule
      \rowcolor{gray!10}
      \textbf{Dataset} & \textbf{\#Correct} & \textbf{\#Incorrect} & \textbf{Positive Ratio (\%)} \\
      \midrule
      BIRD    & 812  & 983  & 54.76 \\
      Spider & 1019 & 532  & 34.30 \\
      EHRSQL   & 141  & 579  & 80.41 \\
      \bottomrule
    \end{tabular}
  \end{minipage}%
  \hfill
  \begin{minipage}[c]{0.45\textwidth}
    \centering
    \includegraphics[width=0.92\textwidth]{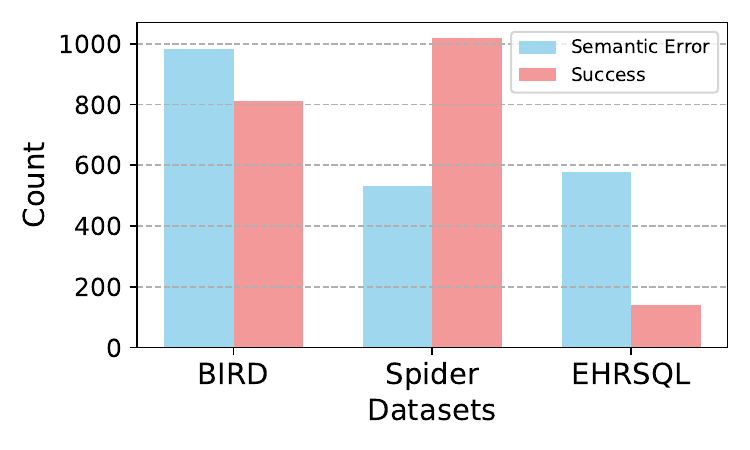}
    \caption{\magentafont{Visualization of class distribution of test sets across datasets.}}
    \label{fig:testset_distribution}
  \end{minipage}
\end{figure}

\magentafont{From the Table~\ref{tab:class_distribution} and the visualization in Figure~\ref{fig:testset_distribution}, we observe a clear difference in the proportion of positive samples (incorrect SQL queries) across datasets: 54.76\% for \textit{BIRD}, 80.41\% for \textit{EHRSQL}, and only 34.30\% for \textit{Spider}. This indicates that \textit{Spider} dataset contains not only fewer incorrect samples but also more subtle ones, making it harder for the model to accurately distinguish these rare and complex errors. As a result, in our experiments, \textit{Spider} tends to show lower AUPRC compared to AUROC, while \textit{BIRD} and \textit{EHRSQL} having a higher proportion of incorrect samples are more likely to achieve higher AUPRC.}

\subsection{Experiment Configuration}
\label{sec:exp-config}

% In our experiments, all approaches are implemented using PyTorch 2.8.0, PyTorch Geometric 2.6.1, Transformers 4.56.1 and CUDA 12.4.
% All experiments are done on a machine equipped with 8 NVIDIA A100 with 80GB memory.

% During model training process, model parameters are optimized by the $\mathrm{AdamW}$ optimizer~\citep{loshchilov2017decoupled}, with a learning rate of $1e-4$ and weight decay of $1e-4$. 
% And the training batch size is set to $32$, dropout rate is set to $0.3$.
% Furthermore, an early-stopping strategy monitoring AUROC metric with a patience of $5$ epochs is employed to mitigate overfitting.

\orangefont{
All experiments were conducted in a Python 3.12 environment. For neural network training and inference, we utilized PyTorch 2.8.0, PyTorch Geometric 2.6.1, Pytorch Scatter 2.1.2, Transformers 4.56.1, and Numpy 1.26.4 SQL preprocessing was performed using SQLGlot 27.7.0 to assist. And the CUDA version of our machine is 12.4. 
To generate the logical plans for SQL queries, we selected Apache Calcite~\citep{begoli2018apache} as the query optimizer due to its extensive support for SQL parsing and optimization. This component was implemented in Java with JDK 21.0.8, utilizing Apache Calcite 1.40.0 as the core optimization library, sqlite-jdbc 3.50.3.0 for database connectivity, and Spring Boot 3.5.4 to enable web service integration and testing with Python environments, with all dependencies managed via Maven.
Finally, all LLM-based baselines are accelerated with vLLM with version 0.9.1 during generation process.
Please refer to our repository and che
}
Our computational environment consists of a server equipped with 8 NVIDIA A100 GPUs, each with 80GB memory, ensuring sufficient resources for large-scale model training and experimentation. 
\orangefont{
To enhance reproducibility, we fix random seeds with $2025$ for all relevant libraries in each experiments.
}

We implement and evaluate each approach using the same experimental pipeline for a fair comparison. Unless otherwise specified, model parameters are optimized by the $\mathrm{AdamW}$ optimizer~\citep{loshchilov2017decoupled}, with an initial learning rate of $1e$-$4$ and weight decay of $1e$-$4$. The training batch size is set to $32$, and dropout is applied with a rate of $0.3$ to mitigate overfitting. Furthermore, an early-stopping strategy is employed by monitoring the validation AUROC metric, with a patience threshold of $5$ epochs.
For model selection, the best checkpoint is chosen based on the highest AUROC achieved in validation dataset during training.

\subsection{Evaluation Metrics for Text-to-SQL Semantic Validation}

In evaluating our Text-to-SQL semantic validation task, a binary classification problem that predicts the correctness of SQL predictions, we adopt two widely used metrics: \textit{Area Under the Precision-Recall Curve} (AUPRC) and \textit{Area Under the Receiver Operating Characteristic Curve} (AUROC).

AUPRC measures the area under the curve that plots precision against recall for different threshold settings. This metric is particularly informative for imbalanced datasets, as it focuses on the model's ability to correctly identify positive examples and ignore the numerous negatives. In our context, a higher AUPRC indicates a better trade-off between precision and recall when distinguishing correct from incorrect SQL queries.

AUROC, on the other hand, represents the area under the ROC curve, which plots the true positive rate (sensitivity) against the false positive rate (1-specificity) across thresholds. AUROC provides an aggregate measure of performance across all classification thresholds, indicating the probability that the model ranks a randomly chosen positive instance higher than a negative one. 

Both AUPRC and AUROC give us a comprehensive view of model’s discriminative ability. 
As our primary objective is to detect semantic errors during the Text-to-SQL process, we \textbf{define invalid question-SQL pairs as positive samples, while valid pairs are treated as negative samples}.

% \begingroup
% \color{orange}
\subsection{Sensitivity Analysis of Query Optimizer for Locial Plan Generation}

\methodname relies on logical plans (LPs) and abstract syntax trees (ASTs) as core foundational components. 
Notably, the method exhibits \textbf{weak sensitivity to the specific choice of query optimizer (aka. LP generator) or its optimization configurations}. 
Modern SQL query optimizers are inherently designed to retain the original query semantics, with their differences primarily manifesting in local decision-making processes—such as join order selection, predicate pushdown strategies, or the insertion of intermediate operators. These variations only induce minor structural discrepancies in the generated LPs and do not alter the high-level semantic information that underpins the validation mechanism of our framework.

Regarding the handling of logical plan extraction failures, these cases should not be considered as semantically incorrect. 
Instances where an LP cannot be successfully generated are almost exclusively attributed to the SQL query failing to pass the database compiler (e.g., due to syntax errors, invalid table/column references, or malformed clauses). In such scenarios, the optimizer abstains from producing an LP entirely, and the database engine returns a clear compilation error. We categorize these cases as \textbf{syntax-level errors rather than semantic-level issues}. As depicted in Figure 1, the error messages generated during compilation are fed back into the iterative query regeneration pipeline with the large language model (LLM), which refines the SQL until it compiles successfully. Only after passing this compilation check does the query proceed to the semantic validation phase of \methodname.

% \endgroup

%% file: 8_4_App_algo.tex
\section{Algorithms for \methodname}
\label{sec: algorithm} 

In this section, we detail the algorithm of context guided property embedding in the hierarchical intermediate representation of the SQL query in \methodname, which is shown in Algorithm~\ref{alg:hero_sql}. 
Besides, we also provide the workflow of the nested message passing process in Algorithm~\ref{alg:nmpnn}.

\begin{algorithm}[H]
    \caption{Hierarchical SQL Representation of \methodname}
    \label{alg:hero_sql}
    \begin{algorithmic}[1]
    \Require Training dataset $\mathcal{D}_{\text{train}} = \{(q_i, s_i)\}$, LLM-based embedding model $g$
    
            % \State Tokenize question $q_i$ to get token sequence $x^q = (x_1^q, \dots, x_{n_q}^q)$
            \For{each logical plan node $v_i^{\mathcal{L}} \in V_{\mathcal{L}}$}
                \For{each AST node $v_j^{\mathcal{A}_i} \in V_i$ of $\mathcal{A}_i$}
                    
                    % \State \textbf{Context-guided Embedding:}
                    \State Fetch AST node embedding with guided context:
                        \[
                        X = [\texttt{CONTEXT}; v_j^{\mathcal{A}_i}]
                        \]
                        
                    \State Compute token embeddings $H = g(X)$
                \EndFor
            \EndFor
    \State \Return hierarchical intermediate representation of SQL query $s_i$
    \end{algorithmic}
\end{algorithm}

\begin{algorithm}[H]
    \caption{Nested Message Passing over AST and LP}
    \label{alg:nmpnn}
    \begin{algorithmic}[1]
    \Require Hierarchical Intermediate Representation of SQL query, AST steps $T_\mathrm{ast}$, LP steps $T_\mathrm{logic}$
    % \Ensure Updated model parameters $\Theta$
    
                \For{each logical plan node $v_i^{\mathcal{L}} \in V_{\mathcal{L}}$}
                    \For{each AST node $v_j^{\mathcal{A}_i} \in V_i$ of $\mathcal{A}_i$}
                        \For{$t = 1$ to $T_\mathrm{ast}$}
                            \State $h_{v_j^{\mathcal{A}_i}}^{(t+1)} \gets 
                                \mathrm{UPDATE}_\mathrm{ast}\Big(
                                    h_{v_j^{\mathcal{A}_i}}^{(t)}, 
                                    \mathrm{AGGREGATE}_\mathrm{ast}(\{h_{v_k^{\mathcal{A}_i}}^{(t)} : v_k^{\mathcal{A}_i} \in \mathcal{N}(v_j^{\mathcal{A}_i})\})
                                \Big)$
                        \EndFor
                    \EndFor
                    \State $h_{v_i^{\mathcal{L}}} \gets \mathrm{POOLING}(\{h_{v_j^{\mathcal{A}_i}}^{(T_\mathrm{ast})}\}_{v_j^{\mathcal{A}_i} \in V_i})$
                \EndFor
            
                \For{$t = 1$ to $T_\mathrm{logic}$}
                    \For{each logical plan node $v_i^{\mathcal{L}} \in V_{\mathcal{L}}$}
                        \State $h_{v_i^{\mathcal{L}}}^{(t+1)} \gets 
                            \mathrm{UPDATE}_\mathrm{logic}\Big(
                                h_{v_i^{\mathcal{L}}}^{(t)}, 
                                \mathrm{AGGREGATE}_\mathrm{logic}(\{h_{v_k^{\mathcal{L}}}^{(t)} : v_k^{\mathcal{L}} \in \mathcal{N}_L(v_i^{\mathcal{L}})\})
                            \Big)$
                    \EndFor
                \EndFor
            
                \State $h_{\mathrm{SQL}} \gets \mathrm{POOLING}(\{h_{v_i^{\mathcal{L}}}^{(T_\mathrm{logic})}\}_{v_i^{\mathcal{L}} \in V_{\mathcal{L}}})$

        % \State \textbf{Fusion and Prediction:}
        %     \State $h_{\mathrm{hadamard}} \leftarrow h_{\mathrm{question}} \odot h_{\mathrm{SQL}}$
        %     \State $\mathbf{h} \leftarrow [h_{\mathrm{question}}; h_{\mathrm{SQL}}; h_{\mathrm{hadamard}}]$
        %     \State $\hat{y}_i \leftarrow \mathrm{MLP}(\mathbf{h})$
        
        % \State \textbf{Loss and Gradient Update:}
        %     \State Compute BCE loss $\mathcal{L}_{\mathrm{BCE}}(\hat{y}_i, y_i)$
        %     \State Backpropagate $\frac{\partial \mathcal{L}_{\mathrm{BCE}}}{\partial \Theta}$ and update $\Theta \leftarrow \Theta - \eta \cdot \frac{\partial \mathcal{L}_{\mathrm{BCE}}}{\partial \Theta}$
    % \EndFor
    \State \Return embedding of SQL query $h_{\mathrm{SQL}}$
    \end{algorithmic}
\end{algorithm}

%% file: 8_8_sen.tex
\section{Sensitivity Analysis}
\label{Sensitivity Analysis}

\subsection{Impact of Number of AST-based Augmented Samples}
\label{sec:ab-neg}

To investigate how performance scales with the number of semantically incorrect samples generated through AST-driven Sub-SQL augmentation, 
we conducted experiments using \methodname with the \textit{embeddinggemma-300m} backbone, varying the number of AST-based augmentation samples included in the training set.
Specifically, we trained models under different negative-to-positive (N/P) ratios: 0.55 (no AST-based augmentation), 0.8, 1.0, and 1.2, and evaluated performance on the \textit{BIRD}, \textit{}{Spider}, and \textit{EHRSQL} datasets. The results are shown in Table~\ref{tab:ratio}.

% \begin{table}[htbp]
% \color{violet}
% \centering
% \caption{
% Sensitivity analysis of different negative-to-positive (N/P) ratios for training dataset.
% }
% \label{tab:ratio}
% \begin{tabular}{c|cc|cc|cc}
% \hline
%  & \multicolumn{2}{c|}{\textbf{BIRD}} & \multicolumn{2}{c|}{\textbf{Spider}} & \multicolumn{2}{c}{\textbf{EHRSQL}} \\
% \textbf{N/P Ratio} & \textbf{AUPRC} & \textbf{AUROC} & \textbf{AUPRC} & \textbf{AUROC} & \textbf{AUPRC} & \textbf{AUROC} \\
% \hline
% 0.55 & 64.39 & 60.56 & 49.73 & 66.19 & 84.14 & 60.77 \\
% 0.8 & \textbf{68.20} & \textbf{63.54} & \underline{50.60} & \underline{66.55} & \underline{85.30} & \underline{63.55} \\
% 1.0 & \underline{67.10} & \underline{63.48} & \textbf{51.11} & \textbf{69.26} & \textbf{85.48} & \textbf{65.92} \\
% 1.2 & 65.24 & 61.44 & 48.35 & 65.74 & 84.52 & 56.19 \\
% \hline
% \end{tabular}

% \end{table}

\begin{wraptable}{r}{0.6\textwidth}
\vspace{-0.4cm}
\centering
% \color{violet}
\caption{
Sensitivity analysis of different negative-to-positive (N/P) ratios for training dataset.
}
\label{tab:ratio}
\resizebox{0.6\textwidth}{!}{
\begin{tabular}{c|cc|cc|cc}
\hline
& \multicolumn{2}{c|}{\textbf{BIRD}} & \multicolumn{2}{c|}{\textbf{Spider}} & \multicolumn{2}{c}{\textbf{EHRSQL}} \\
\textbf{N/P Ratio} & \textbf{AUPRC} & \textbf{AUROC} & \textbf{AUPRC} & \textbf{AUROC} & \textbf{AUPRC} & \textbf{AUROC} \\
\hline
0.55 & 64.39 & 60.56 & 49.73 & 66.19 & 84.14 & 60.77 \\
0.8 & \textbf{68.20} & \textbf{63.54} & \underline{50.60} & \underline{66.55} & \underline{85.30} & \underline{63.55} \\
1.0 & \underline{67.10} & \underline{63.48} & \textbf{51.11} & \textbf{69.26} & \textbf{85.48} & \textbf{65.92} \\
1.2 & 65.24 & 61.44 & 48.35 & 65.74 & 84.52 & 56.19 \\
\hline
\end{tabular}
}
\vspace{-0.4cm}
\end{wraptable}

% which indicate that maintaining a balanced and trade-off negative-to-positive ratio of approximately 1:1 is an effective and robust strategy. This ratio provides sufficient semantic contrast without overwhelming the model with noisy or overly abundant negative samples, ultimately supporting better generalization and more stable learning dynamics.

Our findings show that increasing the N/P ratio from 0.55 → 0.8 → 1.0 consistently improves performance for most datasets. The best results for \textit{Spider} and \textit{EHRSQL} appear at an N/P ratio of 1.0, while \textit{BIRD}’s best is at 0.8, with only a marginal difference at 1.0, which corresponds to prior research's conclusions~\citep{robinson2020contrastive, gao2021simcse, wu2022esimcse, chawla2002smote}.
However, increasing the ratio to 1.2 leads to a noticeable performance drop across all datasets, suggesting that an excessive number of negative samples may introduce noise, reduce signal-to-noise ratio, or skew the data distribution in a way that impairs learning.

Overall, these results indicate that \textbf{maintaining a balanced and trade-off negative-to-positive ratio of approximately 1:1 is an effective and robust strategy}. This ratio provides sufficient semantic contrast without overwhelming the model with noisy or overly abundant negative samples, ultimately supporting better generalization and more stable learning dynamics.

\subsection{Impact of layer numbers in NMPNN}

\begin{wrapfigure}{r}{9cm}
    \vspace{-0.5cm}
    \raggedright % align to the right
    \includegraphics[scale=0.21]{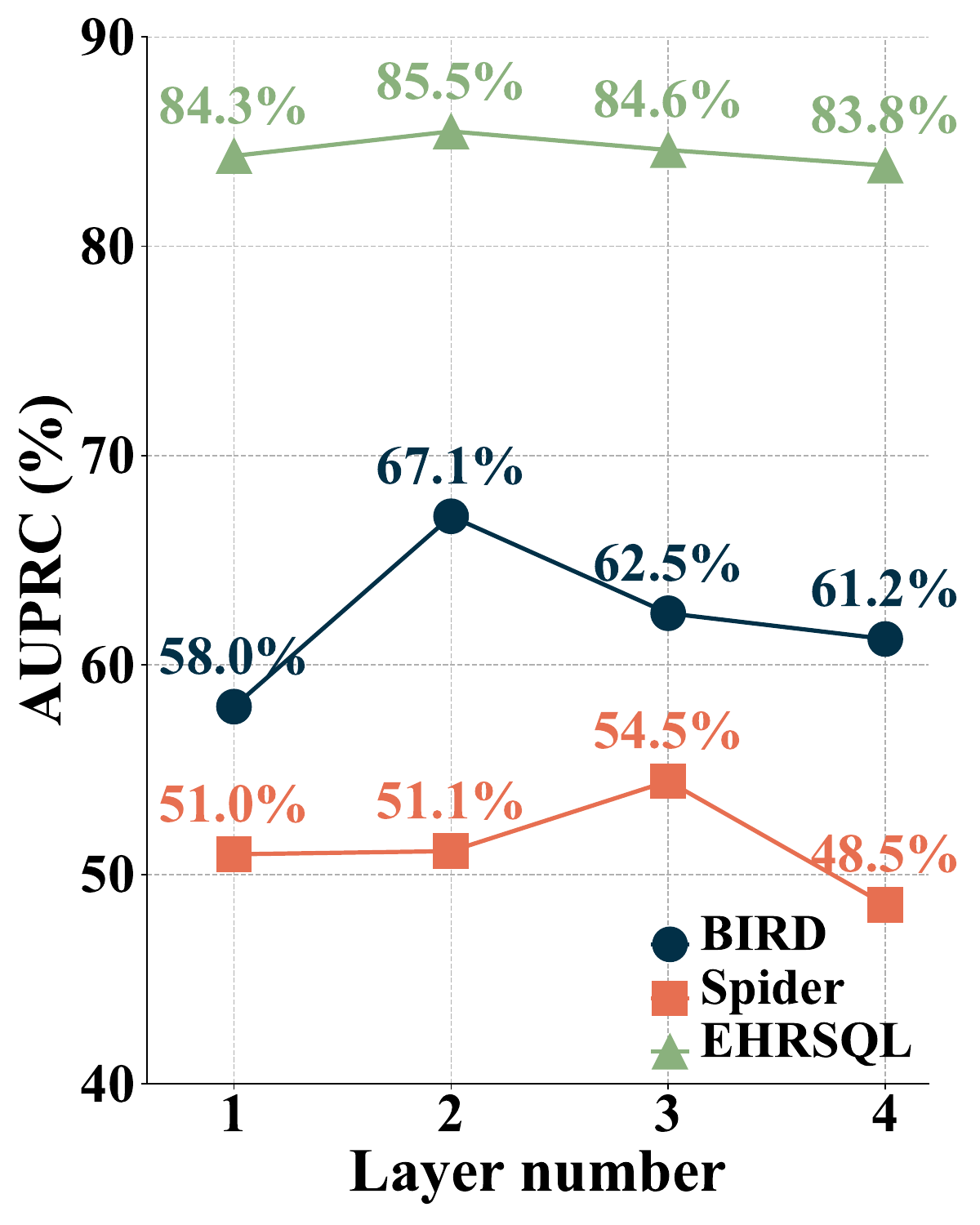}
    \includegraphics[scale=0.21]{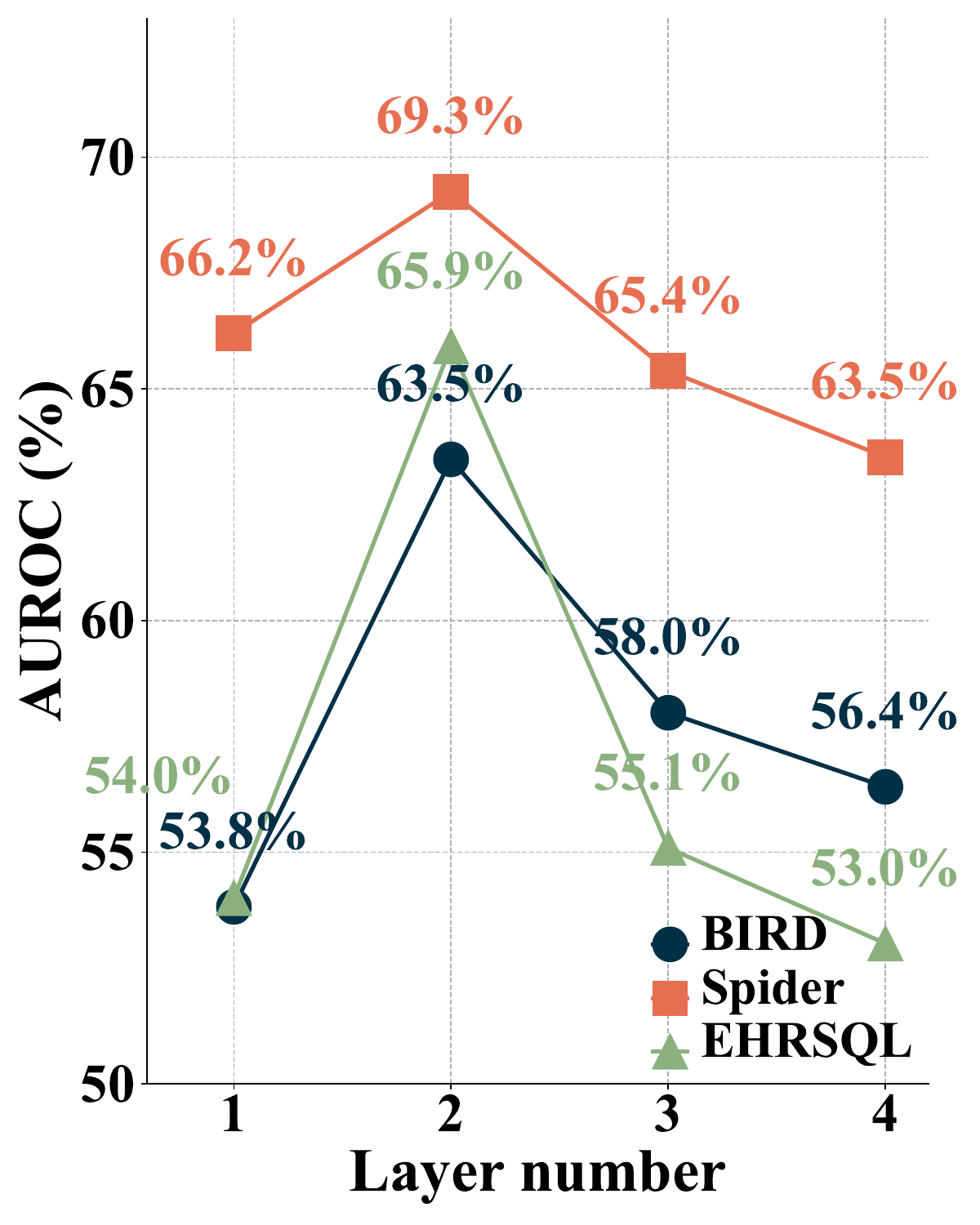}
    \caption{Sensitivity analysis of layer number $l$ in nested message passing neural network}
    \label{fig:proto_sensitivity}
    \vspace{-0.4cm}
\end{wrapfigure}

We examine the impact of layer numbers in the NMPNN of \methodname.
We specifically analyze the variations of AUPRC and AUROC for different layer numbers $l$ from the list $[1, 2, 3, 4]$ with the embedding model \textit{embeddinggemma-300m}.
As shown in Figure~\ref{fig:proto_sensitivity}, the performance reaches its peak when the layer number is set to 2 across almost all evaluation metrics. When the layer number exceeds 2, the model’s performance degrades, likely due to increased training difficulty and the over-smoothing effect, where node representations become indistinguishable and fail to capture fine-grained structural details needed for effective semantic validation. Conversely, with only one layer, the model struggles to sufficiently aggregate structural information from both the LP and AST, which prevents it from fully modeling the relationships among different nodes in SQL queries. 
Therefore, a layer number of 2 provides the optimal balance between information propagation and feature discrimination.

%% file: 8_5_App_more.tex
\section{Additional Experiments}

% \begingroup
% \color{teal}
\subsection{Performance on More Challenging Dataset}

To further verify the effectiveness of our method, we evaluate
\methodname and baseline methods on more challenging datasets like \textit{Spider 2.0}~\citep{leispider}. 
\textit{Spider 2.0} dataset is a quite challenging dataset whose questions are very difficult, schemas are much longer than other datasets, and requires complex SQL queries that may including nested SQL clauses, common table expression (CTE), and window functions to deal with them.

The results are shown shown in Table~\ref{tab:spider_bird_ehrsql}, \methodname achieves the best performance across all configurations on \textit{Spider 2.0} dataset, outperforming all baselines and demonstrating strong robustness in handling complex SQL queries. 
These empirical results provide clear evidence that \methodname scales effectively to more complex and realistic SQL scenarios.

Actually, \methodname is designed with \textbf{sufficient generality to process any syntactically correct SQL statement} that can be compiled by the corresponding SQL engine, including those containing advanced constructs. 
As long as the SQL statement is syntactically valid, we can obtain its logical plan (LP) and abstract syntax tree (AST), upon which \methodname performs semantic validation.
In Table~\ref{Case_HEROSQL_Generality}, we present a complex SQL query with CTE Structures and show that \methodname can handle it well and reflect the error part.

\begin{table*}[htbp]
% \color{teal}
\centering
% \caption{Illustration of \methodname's capability to process complex SQL with CTEs and nested structures.}
\caption{
Case Study on Text-to-SQL Error Detection with \methodname in Handling Complex SQL with CTE Structures
}

\label{Case_HEROSQL_Generality}
\begin{tabular}{
  >{\centering\arraybackslash}m{2.5cm}
  >{\raggedright\arraybackslash}m{11cm}
}
\hline
\textbf{Description} & \methodname is designed with sufficient generality to process any syntactically correct SQL statement that can be compiled by the underlying SQL engine, including advanced constructs such as nested queries, common table expressions (CTEs), and window functions. As long as a query is syntactically valid, the system retrieves its logical plan (LP) and abstract syntax tree (AST), upon which \methodname performs semantic validation. \\ \hline

\textbf{Example SQL} &
\begin{tcolorbox}[colback=blue!12!white,colframe=white,boxsep=1pt,left=2pt,right=2pt,top=2pt,bottom=2pt,after=\vspace{-2pt}]
[Question] \\
List the employees in the company whose salaries are higher than 100{,}000, as well as the average salary of their respective departments. \\[2pt]

[SQL] \\
WITH DeptAvg AS ( \\
\quad SELECT Department, AVG(Salary) AS AvgSalary \\
\quad FROM Employees \\
\quad GROUP BY Department \\
), \\
HighEarners AS ( \\
\quad SELECT Name, Department, Salary \\
\quad FROM Employees \\
\quad WHERE Salary $>$ 100000 \\
) \\
SELECT h.Name, h.Salary, d.AvgSalary \\
FROM HighEarners h \\
JOIN DeptAvg d ON h.Department = d.Department;
\end{tcolorbox}
\\ \hline

\textbf{Logical Plan} &
\begin{tcolorbox}[colback=green!12!white,colframe=white,boxsep=1pt,left=2pt,right=2pt,top=2pt,bottom=2pt,after=\vspace{-2pt}]
Project(Name=[$0$], Salary=[$2$], AvgSalary=[$5$]) \\
\quad Join(condition=[=($1$, $4$)], joinType=[inner]) \\
\qquad Project(Name=[$0$], Department=[$1$], Salary=[$2$]) \\
\qquad\quad Filter(condition=[$\ge$($2$, 100000)]) \\
\qquad\qquad TableScan(table=[[Employees]]) \\[2pt]
\qquad Aggregate(group=[\{Department\}], AvgSalary=[AVG($2$)]) \\
\qquad\quad Project(Department=[$1$], Salary=[$2$]) \\
\qquad\qquad TableScan(table=[[Employees]])
\end{tcolorbox}
\\ \hline

\textbf{Explanation} &
\begin{tcolorbox}[colback=yellow!15!white,colframe=white,boxsep=1pt,left=2pt,right=2pt,top=2pt,bottom=2pt,after=\vspace{-2pt}]
The logical plan decomposes the two CTEs into standard relational operators, explicitly exposing the global semantics of the query. The first branch filters employees with salaries above 100{,}000, while the second computes the average salary for each department. The final join aligns high earners with the corresponding departmental averages. Operating on this structured LP/AST representation, \methodname robustly handles compositional and deeply nested SQL queries. \\[4pt]
More broadly, logical-plan–based reasoning enables \methodname to capture the global intent and hierarchical organization inherent in sophisticated SQL. This design supports principled interpretation of nested SQL, CTE expansions, and window-function semantics, which encode analytical operations that are difficult to interpret reliably at the surface-syntax level.
\end{tcolorbox}
\\ \hline

\end{tabular}
\end{table*}

% \endgroup

% \begingroup
% \color{olive}
\subsection{Performance on Ambiguous Text-to-SQL Dataset}

\begin{wraptable}{r}{0.6\textwidth}
\vspace{-0.45cm}
% \begin{table}[ht]
% \color{olive}
\centering
\caption{
    Performance comparison of different methods on the \textit{Ambrosia} benchmark.
}
\label{tab:ambrosia-comparison}
% \resizebox{\linewidth}{!}{
\renewcommand{\arraystretch}{1.2}
\begin{tabular}{lcccc}
\hline
\multirow{2}{*}{\textbf{Method}}
    & \multicolumn{2}{c}{\textbf{Qwen3-0.6b}}
    & \multicolumn{2}{c}{\textbf{Gemma-3-0.3b}} \\
\cline{2-5}
 & \textbf{AUPRC} & \textbf{AUROC}
 & \textbf{AUPRC} & \textbf{AUROC} \\
\hline
prompt      & 58.18 & 48.75 & 58.36 & 50.00 \\
CoT         & 59.04 & 49.79 & 58.38 & 50.04 \\
ConfScore   & 59.61 & 50.63 & 59.58 & 51.56 \\
COVE        & 57.88 & 48.80 & 58.64 & 50.58 \\
TED         & 63.91 & 56.95 & 60.51 & 50.70 \\
\midrule
\methodname     & \textbf{64.18} & \textbf{57.28} & \textbf{61.72} & \textbf{52.06} \\
\hline
\end{tabular}
% }
% \end{table}
\vspace{-0.4cm}
\end{wraptable}

% \textit{Ambrosia}~\citep{saparina2024ambrosia} is a cross-domain benchmark dedicated to parsing \textbf{ambiguous} natural-language questions into SQL. It contains 1,277 ambiguous questions, 2,965 corresponding SQL interpretations, and 846 multi-table databases spanning 16 realistic domains (e.g., Banking, Entertainment, Healthcare). Each ambiguous question is paired with all valid, human-verified SQL queries that arise from scope, attachment, or vagueness ambiguities, yielding 2–3 gold queries per question. The dataset is split 90 \% for zero shot evaluation and 10 \% for few-shot demonstration; questions are grouped by ambiguity type rather than by database to preserve linguistic variety. Compared with prior text-to-SQL corpora, Ambrosia introduces the first large-scale testbed where ambiguity persists even when the full schema and content are known, requiring models to explicitly recognize and generate multiple correct interpretations instead of a single canonical query.

To evaluate the robustness of \methodname in real-world scenarios, we conducted experiments on the \textit{Ambrosia}~\citep{saparina2024ambrosia} dataset. 
\textit{Ambrosia} contains naturally ambiguous questions, each paired with 2–3 correct SQL interpretations, designed to test the generalization ability of models under ambiguity and complex schema information. The experimental results on \textit{Ambrosia} dataset are reported in Table~\ref{tab:ambrosia-comparison}.

Across all settings, \methodname maintains strong generalization ability  on ambiguous and noisy-schema queries. These results indicate that \methodname is resilient to noise and schema-level ambiguity. This suggests that \methodname can generalize well even in real-world environments where schema noise and ambiguity are common.

% \endgroup

\subsection{Latency and Throughput for Interactive Validation}

% \begingroup
% \color{brown}

To understand the computational cost of our method, we conduct experiments measuring end-to-end inference latency and throughput, comparing our method \methodname with the pure LLM baseline Prompt, which is the lightest method among all baselines. 
The test are conducted on 4 RTX 3090 GPUs using \textit{Qwen3-0.6B} as the LLM backbone on BIRD dataset. The results are summarized in Table~\ref{tab:latency}.

\begin{table}[ht]
% \color{brown}
\centering
\caption{Comparison of throughput and average latency between pure LLM method and \methodname.}
\label{tab:throughput_latency}
\label{tab:latency}
\begin{tabular}{lcc}
\hline
\textbf{Method} & \textbf{Throughput (samples/s)} & \textbf{Avg. Latency (ms/sample)} \\
\hline
Prompt & 130.21 & 7.68 \\
\methodname  & 85.47 & 11.70 \\
\hline
\end{tabular}
\end{table}

From the result in Table~\ref{tab:latency}, we can see that \textbf{our approach maintains comparable throughput and latency to the Prompt-based method}, without introducing significant delays or reducing overall performance.
This is because that the efficiency of our approach mainly stems from the lightweight two-layer \textit{Nested Message Passing Neural Network} (NMPNN) backbone, which consists of GAT layers and small MLPs. The average per-sample latency (11.70 ms) is just slightly higher than that of the baseline (7.68 ms).

Overall, these results demonstrate that \textbf{\methodname provides competitive latency and throughput for interactive usage}. 
Having an average end-to-end per-query latency of 11.7 ms remains well within typical real-time validation constraints.

% \endgroup

\tealfont{
On scalability, we fully recognize the importance of handling long contexts and complex schemas in real-world enterprise scenarios. Our framework is deliberately designed to be model-agnostic and compatible with any scalable backbone LLM and message-passing architecture. This allows \methodname to naturally benefit from advances in scalable GNNs and graph training techniques, such as LMC~\citep{shi2023lmc}, Sketch-GNN~\citep{ding2022sketch}, REST~\citep{xue2024haste} MeGraph~\citep{dong2023megraph}, and Ginex~\citep{park2022vldb}, as well as from future improvements in large-context LLMs.
}

\tealfont{
For long-context scalability specifically, by adopting a decoder-only pretrained LLM as the embedding backbone, our method inherits the context window and scalability properties of state-of-the-art LLMs. This enables us to process very large and complex database schemas and SQL queries, on par with or beyond existing LLM-based Text-to-SQL validators.
}

% \begingroup
% \color{magenta}
\subsection{Performance Comparison Between \methodname and Commercial Models}

To evaluate the performance difference between our approach and directly using prompts with commercial models for SQL semantic validation, we conducted experiments with strong commercial LLM backbones, including \textit{GPT-4o}, \textit{GPT-o4-mini}, and \textit{Claude-haiku-4-5}. These models are used in a prompt-only setting, where the LLM directly judges SQL semantic correctness without any additional training.

For \methodname, we further evaluated different Qwen3 backbones, including Qwen3-0.6B and Qwen3-4B, to examine whether \methodname continues to provide improvements when paired with larger open-source models. The results are summarized in Table~\cref{tab:commercial}.

% \begin{table}[ht]
% \color{magenta}
% \centering
% \caption{
%     Comparison of the performance of different models.
% }
% \label{tab:commercial}
% \resizebox{\linewidth}{!}{
% \renewcommand{\arraystretch}{1.2}
% \begin{tabular}{lccccccccc}
% \hline
% \multirow{2}{*}{\textbf{Method}}
%     & \multirow{2}{*}{\textbf{Base}}
%     & \multicolumn{2}{c}{\textbf{BIRD}} 
%     & \multicolumn{2}{c}{\textbf{Spider}} 
%     & \multicolumn{2}{c}{\textbf{EHRSQL}}
%     & \multicolumn{2}{c}{\textbf{Spider 2.0}} \\
% \cline{3-10}
%  &  & \textbf{AUPRC} & \textbf{AUROC} 
%     & \textbf{AUPRC} & \textbf{AUROC} 
%     & \textbf{AUPRC} & \textbf{AUROC} 
%     & \textbf{AUPRC} & \textbf{AUROC} \\
% \hline

% \multirow{4}{*}{\textbf{Prompt}} 
%     & Qwen3-0.6b           & 60.85 & 57.25 & 40.86 & 58.94 & 84.72 & 58.86 & 88.78 & 51.80 \\
%     & GPT-4o               & 73.56 & 73.45 & 47.68 & 66.37 & 86.78 & 67.04 & - & - \\
%     & GPT-o4-mini          & \textbf{80.36} & \textbf{79.25} & \textbf{63.21} & \textbf{75.81} & 86.38 & 68.00 & - & - \\
%     & Claude-haiku-4-5     & 75.87 & 76.15 & 48.72 & 65.80 & 83.92 & 61.26 & - & - \\
% \hline

% \multirow{2}{*}{\methodname} 
%     & Qwen3-0.6b           & 67.39 & 61.51 & 51.92 & 67.32 & 89.07 & 69.53 & 92.59 & 64.32 \\
%     & Qwen3-4b             & 71.18 & 64.17 & 53.52 & 69.78 & \textbf{89.28} & \textbf{72.03} & - & - \\
% \hline

% \end{tabular}
% }
% \end{table}

\begin{table}[ht]
% \color{magenta}
\centering
\caption{
    Comparison of the performance of different models.
}
\label{tab:commercial}
\resizebox{\linewidth}{!}{
\renewcommand{\arraystretch}{1.2}
\begin{tabular}{lccccccc}
\hline
\multirow{2}{*}{\textbf{Method}}
    & \multirow{2}{*}{\textbf{Base}}
    & \multicolumn{2}{c}{\textbf{BIRD}} 
    & \multicolumn{2}{c}{\textbf{Spider}} 
    & \multicolumn{2}{c}{\textbf{EHRSQL}}\\
\cline{3-8}
 &  & \textbf{AUPRC} & \textbf{AUROC} 
    & \textbf{AUPRC} & \textbf{AUROC} 
    & \textbf{AUPRC} & \textbf{AUROC}  \\
\hline

\multirow{4}{*}{Prompt} 
    & Qwen3-0.6b           & 60.85 & 57.25 & 40.86 & 58.94 & 84.72 & 58.86 \\
    & GPT-4o               & 73.56 & 73.45 & 47.68 & 66.37 & 86.78 & 67.04 \\
    & GPT-o4-mini          & \textbf{80.36} & \textbf{79.25} & \textbf{63.21} & \textbf{75.81} & 86.38 & 68.00\\
    & Claude-haiku-4-5     & 75.87 & 76.15 & 48.72 & 65.80 & 83.92 & 61.26 \\
\hline

\multirow{2}{*}{\methodname} 
    & Qwen3-0.6b           & 67.39 & 61.51 & 51.92 & 67.32 & 89.07 & 69.53  \\
    & Qwen3-4b             & 71.18 & 64.17 & 53.52 & 69.78 & \textbf{89.28} & \textbf{72.03}\\
\hline

\end{tabular}
}
\end{table}

From these results, we have the following observations.
\ding{182} \textbf{Model capability correlates with prompt-only baseline performance}: stronger commercial LLMs (e.g., GPT-o4-mini, GPT-4o, Claude-haiku-4-5) outperform lightweight open-source prompt baselines on all datasets.
\ding{183} \textbf{Within the Qwen family, scaling up improves performance}: \methodname with \textit{Qwen3-4B} consistently outperforms \methodname with \textit{Qwen3-0.6B}, confirming that our framework naturally benefits from more capable backbones.
\ding{184} \methodname adds value on top of a given backbone: even when starting from a relatively small open-source model \textit{Qwen3-0.6B}, \methodname achieves substantial gains over its prompt-only counterpart, and with Qwen3-4B it reaches performance that is competitive with or better than strong commercial prompt-only baselines on several metrics.

It is worth noting that our contribution is a framework for hierarchical SQL semantic validation, rather than a new pretrained model. 
\methodname is model-agnostic: its capability depends on the underlying pretrained LLM-based embedding model and is compatible with a wide range of pretrained models, from lightweight open-source models to state-of-the-art commercial LLMs. 

The main experiments are conducted primarily with two small open-source encoders \textit{Qwen3-0.6B} and \textit{Gemma-3-0.3B} to demonstrate that the gains come from the framework itself, not merely from model scale.

When selecting LLM backbones, we chose open-source models instead of commercial ones mainly for three practical reasons:

\begin{itemize}[leftmargin=*]
    \item \textbf{Privacy}. Real-world database applications often involve sensitive or proprietary data. In many enterprise settings, only local offline deployment is allowed, and sending data to external commercial APIs to call these commercial but powerful models is not acceptable.
    \item \textbf{Cost}. Training \methodname requires repeatedly invoking embedding models over large augmented datasets. Using commercial APIs as embedding backbones would incur prohibitive costs.
    \item \textbf{Efficiency}. Frequent remote API calls for embeddings would also significantly slow down training and experimentation, making the framework impractical for real deployment and iteration.
\end{itemize}

% \endgroup

%% file: iclr2026_conference.bib
@misc{liu2025surveytexttosqlerallms,
      title={A Survey of Text-to-SQL in the Era of LLMs: Where are we, and where are we going?}, 
      author={Xinyu Liu and Shuyu Shen and Boyan Li and Peixian Ma and Runzhi Jiang and Yuxin Zhang and Ju Fan and Guoliang Li and Nan Tang and Yuyu Luo},
      year={2025},
      eprint={2408.05109},
      archivePrefix={arXiv},
      primaryClass={cs.DB},
      url={https://arxiv.org/abs/2408.05109}, 
}

@article{liu2025survey,
  title={A Survey of Text-to-SQL in the Era of LLMs: Where are we, and where are we going?},
  author={Liu, Xinyu and Shen, Shuyu and Li, Boyan and Ma, Peixian and Jiang, Runzhi and Zhang, Yuxin and Fan, Ju and Li, Guoliang and Tang, Nan and Luo, Yuyu},
  journal={IEEE Transactions on Knowledge and Data Engineering},
  year={2025},
  publisher={IEEE}
}

@article{shi2024survey,
  title={A survey on employing large language models for text-to-sql tasks},
  author={Shi, Liang and Tang, Zhengju and Zhang, Nan and Zhang, Xiaotong and Yang, Zhi},
  journal={ACM Computing Surveys},
  year={2024},
  publisher={ACM New York, NY}
}

@inproceedings{begoli2018apache,
  title={Apache calcite: A foundational framework for optimized query processing over heterogeneous data sources},
  author={Begoli, Edmon and Camacho-Rodr{\'\i}guez, Jes{\'u}s and Hyde, Julian and Mior, Michael J and Lemire, Daniel},
  booktitle={Proceedings of the 2018 International Conference on Management of Data},
  pages={221--230},
  year={2018}
}

@inproceedings{soliman2014orca,
  title={Orca: a modular query optimizer architecture for big data},
  author={Soliman, Mohamed A and Antova, Lyublena and Raghavan, Venkatesh and El-Helw, Amr and Gu, Zhongxian and Shen, Entong and Caragea, George C and Garcia-Alvarado, Carlos and Rahman, Foyzur and Petropoulos, Michalis and others},
  booktitle={Proceedings of the 2014 ACM SIGMOD international conference on Management of data},
  pages={337--348},
  year={2014}
}

@inproceedings{somov2025confidence,
  title={Confidence estimation for error detection in text-to-sql systems},
  author={Somov, Oleg and Tutubalina, Elena},
  booktitle={Proceedings of the AAAI Conference on Artificial Intelligence},
  volume={39},
  pages={25137--25145},
  year={2025}
}

@inproceedings{askari2025magic,
  title={Magic: Generating self-correction guideline for in-context text-to-sql},
  author={Askari, Arian and Poelitz, Christian and Tang, Xinye},
  booktitle={Proceedings of the AAAI Conference on Artificial Intelligence},
  volume={39},
  pages={23433--23441},
  year={2025}
}

@inproceedings{chen-etal-2023-error,
    title = "Error Detection for Text-to-{SQL} Semantic Parsing",
    author = "Shijie Chen and Ziru Chen and Huan Sun and Yu Su",
    booktitle = "Findings of the Association for Computational Linguistics: EMNLP 2023",
    month = dec,
    year = "2023",
    address = "Singapore",
    publisher = "Association for Computational Linguistics",
    url = "https://aclanthology.org/2023.findings-emnlp.785",
    doi = "10.18653/v1/2023.findings-emnlp.785",
    pages = "11730--11743",
}

@misc{liu2025nl2sqlbugsbenchmarkdetectingsemantic,
      title={NL2SQL-BUGs: A Benchmark for Detecting Semantic Errors in NL2SQL Translation}, 
      author={Xinyu Liu and Shuyu Shen and Boyan Li and Nan Tang and Yuyu Luo},
      year={2025},
      eprint={2503.11984},
      archivePrefix={arXiv},
      primaryClass={cs.DB},
      url={https://arxiv.org/abs/2503.11984}, 
}

@inproceedings{DBCopilot,
  author       = {Tianshu Wang and
                  Xiaoyang Chen and
                  Hongyu Lin and
                  Xianpei Han and
                  Le Sun and
                  Hao Wang and
                  Zhenyu Zeng},
  editor       = {Alkis Simitsis and
                  Bettina Kemme and
                  Anna Queralt and
                  Oscar Romero and
                  Petar Jovanovic},
  title        = {DBCopilot: Natural Language Querying over Massive Databases via Schema
                  Routing},
  booktitle    = {Proceedings 28th International Conference on Extending Database Technology,
                  {EDBT} 2025, Barcelona, Spain, March 25-28, 2025},
  pages        = {707--721},
  publisher    = {OpenProceedings.org},
  year         = {2025},
  url          = {https://doi.org/10.48786/edbt.2025.57},
  doi          = {10.48786/EDBT.2025.57},
  bibsource    = {dblp computer science bibliography, https://dblp.org}
}

@article{cheng2024snil,
  title={SNIL: generating sports news from insights with large language models},
  author={Cheng, Liqi and Deng, Dazhen and Xie, Xiao and Qiu, Rihong and Xu, Mingliang and Wu, Yingcai},
  journal={IEEE Transactions on Visualization and Computer Graphics},
  year={2024},
  publisher={IEEE}
}

@article{lian2024chatbi,
  title={ChatBI: Towards natural language to complex business intelligence SQL},
  author={Lian, Jinqing and Liu, Xinyi and Shao, Yingxia and Dong, Yang and Wang, Ming and Wei, Zhang and Wan, Tianqi and Dong, Ming and Yan, Hailin},
  journal={arXiv preprint arXiv:2405.00527},
  year={2024}
}

@article{xie2025opensearch,
  title={Opensearch-sql: Enhancing text-to-sql with dynamic few-shot and consistency alignment},
  author={Xie, Xiangjin and Xu, Guangwei and Zhao, Lingyan and Guo, Ruijie},
  journal={Proceedings of the ACM on Management of Data},
  volume={3},
  number={3},
  pages={1--24},
  year={2025},
  publisher={ACM New York, NY, USA}
}

@inproceedings{ICLR2025_CHASE_SQL,
 author = {Pourreza, Mohammadreza and Li, Hailong and Sun, Ruoxi and Chung, Yeounoh and Talaei, Shayan and Kakkar, Gaurav Tarlok and Gan, Yu and Saberi, Amin and Ozcan, Fatma and Arik, Sercan},
 booktitle = {International Conference on Representation Learning},
 editor = {Y. Yue and A. Garg and N. Peng and F. Sha and R. Yu},
 pages = {60385--60415},
 title = {CHASE-SQL: Multi-Path Reasoning and Preference Optimized Candidate Selection in Text-to-SQL},
 url = {https://proceedings.iclr.cc/paper_files/paper/2025/file/974ff7b5bf08dbf9400b5d599a39c77f-Paper-Conference.pdf},
 volume = {2025},
 year = {2025}
}

@article{liao2025learnat,
  title={LearNAT: Learning NL2SQL with AST-guided Task Decomposition for Large Language Models},
  author={Liao, Weibin and Gao, Xin and Jia, Tianyu and Qiu, Rihong and Zhu, Yifan and Lin, Yang and Chu, Xu and Zhao, Junfeng and Wang, Yasha},
  journal={arXiv preprint arXiv:2504.02327},
  year={2025}
}

@article{yang2025qwen3,
  title={Qwen3 technical report},
  author={Yang, An and Li, Anfeng and Yang, Baosong and Zhang, Beichen and Hui, Binyuan and Zheng, Bo and Yu, Bowen and Gao, Chang and Huang, Chengen and Lv, Chenxu and others},
  journal={arXiv preprint arXiv:2505.09388},
  year={2025}
}

@article{zhang2025qwen3,
  title={Qwen3 Embedding: Advancing Text Embedding and Reranking Through Foundation Models},
  author={Zhang, Yanzhao and Li, Mingxin and Long, Dingkun and Zhang, Xin and Lin, Huan and Yang, Baosong and Xie, Pengjun and Yang, An and Liu, Dayiheng and Lin, Junyang and others},
  journal={arXiv preprint arXiv:2506.05176},
  year={2025}
}

@misc{gemmateam2025gemma3technicalreport,
      title={Gemma 3 Technical Report}, 
      author={Gemma Team and Aishwarya Kamath and Johan Ferret and Shreya Pathak and Nino Vieillard and Ramona Merhej and Sarah Perrin and Tatiana Matejovicova and Alexandre Ramé and Morgane Rivière and Louis Rouillard and Thomas Mesnard and Geoffrey Cideron and Jean-bastien Grill and Sabela Ramos and Edouard Yvinec and Michelle Casbon and Etienne Pot and Ivo Penchev and Gaël Liu and Francesco Visin and Kathleen Kenealy and Lucas Beyer and Xiaohai Zhai and Anton Tsitsulin and Robert Busa-Fekete and Alex Feng and Noveen Sachdeva and Benjamin Coleman and Yi Gao and Basil Mustafa and Iain Barr and Emilio Parisotto and David Tian and Matan Eyal and Colin Cherry and Jan-Thorsten Peter and Danila Sinopalnikov and Surya Bhupatiraju and Rishabh Agarwal and Mehran Kazemi and Dan Malkin and Ravin Kumar and David Vilar and Idan Brusilovsky and Jiaming Luo and Andreas Steiner and Abe Friesen and Abhanshu Sharma and Abheesht Sharma and Adi Mayrav Gilady and Adrian Goedeckemeyer and Alaa Saade and Alex Feng and Alexander Kolesnikov and Alexei Bendebury and Alvin Abdagic and Amit Vadi and András György and André Susano Pinto and Anil Das and Ankur Bapna and Antoine Miech and Antoine Yang and Antonia Paterson and Ashish Shenoy and Ayan Chakrabarti and Bilal Piot and Bo Wu and Bobak Shahriari and Bryce Petrini and Charlie Chen and Charline Le Lan and Christopher A. Choquette-Choo and CJ Carey and Cormac Brick and Daniel Deutsch and Danielle Eisenbud and Dee Cattle and Derek Cheng and Dimitris Paparas and Divyashree Shivakumar Sreepathihalli and Doug Reid and Dustin Tran and Dustin Zelle and Eric Noland and Erwin Huizenga and Eugene Kharitonov and Frederick Liu and Gagik Amirkhanyan and Glenn Cameron and Hadi Hashemi and Hanna Klimczak-Plucińska and Harman Singh and Harsh Mehta and Harshal Tushar Lehri and Hussein Hazimeh and Ian Ballantyne and Idan Szpektor and Ivan Nardini and Jean Pouget-Abadie and Jetha Chan and Joe Stanton and John Wieting and Jonathan Lai and Jordi Orbay and Joseph Fernandez and Josh Newlan and Ju-yeong Ji and Jyotinder Singh and Kat Black and Kathy Yu and Kevin Hui and Kiran Vodrahalli and Klaus Greff and Linhai Qiu and Marcella Valentine and Marina Coelho and Marvin Ritter and Matt Hoffman and Matthew Watson and Mayank Chaturvedi and Michael Moynihan and Min Ma and Nabila Babar and Natasha Noy and Nathan Byrd and Nick Roy and Nikola Momchev and Nilay Chauhan and Noveen Sachdeva and Oskar Bunyan and Pankil Botarda and Paul Caron and Paul Kishan Rubenstein and Phil Culliton and Philipp Schmid and Pier Giuseppe Sessa and Pingmei Xu and Piotr Stanczyk and Pouya Tafti and Rakesh Shivanna and Renjie Wu and Renke Pan and Reza Rokni and Rob Willoughby and Rohith Vallu and Ryan Mullins and Sammy Jerome and Sara Smoot and Sertan Girgin and Shariq Iqbal and Shashir Reddy and Shruti Sheth and Siim Põder and Sijal Bhatnagar and Sindhu Raghuram Panyam and Sivan Eiger and Susan Zhang and Tianqi Liu and Trevor Yacovone and Tyler Liechty and Uday Kalra and Utku Evci and Vedant Misra and Vincent Roseberry and Vlad Feinberg and Vlad Kolesnikov and Woohyun Han and Woosuk Kwon and Xi Chen and Yinlam Chow and Yuvein Zhu and Zichuan Wei and Zoltan Egyed and Victor Cotruta and Minh Giang and Phoebe Kirk and Anand Rao and Kat Black and Nabila Babar and Jessica Lo and Erica Moreira and Luiz Gustavo Martins and Omar Sanseviero and Lucas Gonzalez and Zach Gleicher and Tris Warkentin and Vahab Mirrokni and Evan Senter and Eli Collins and Joelle Barral and Zoubin Ghahramani and Raia Hadsell and Yossi Matias and D. Sculley and Slav Petrov and Noah Fiedel and Noam Shazeer and Oriol Vinyals and Jeff Dean and Demis Hassabis and Koray Kavukcuoglu and Clement Farabet and Elena Buchatskaya and Jean-Baptiste Alayrac and Rohan Anil and Dmitry and Lepikhin and Sebastian Borgeaud and Olivier Bachem and Armand Joulin and Alek Andreev and Cassidy Hardin and Robert Dadashi and Léonard Hussenot},
      year={2025},
      eprint={2503.19786},
      archivePrefix={arXiv},
      primaryClass={cs.CL},
      url={https://arxiv.org/abs/2503.19786}, 
}

@inproceedings{devlin2019bert,
  title={Bert: Pre-training of deep bidirectional transformers for language understanding},
  author={Devlin, Jacob and Chang, Ming-Wei and Lee, Kenton and Toutanova, Kristina},
  booktitle={Proceedings of the 2019 conference of the North American chapter of the association for computational linguistics: human language technologies, volume 1 (long and short papers)},
  pages={4171--4186},
  year={2019}
}

@article{raffel2020exploring,
title={Exploring the limits of transfer learning with a unified text-to-text transformer},
author={Raffel, Colin and Shazeer, Noam and Roberts, Adam and et al.},
journal={JMLR},
year={2020},
publisher={JMLRORG}
}

@inproceedings{Feng2020CodeBERT,
  title = {{{CodeBERT}}: {{A Pre-Trained Model}} for {{Programming}} and {{Natural Languages}}},
  shorttitle = {{{CodeBERT}}},
  booktitle = {Findings of the {{Association}} for {{Computational Linguistics}}: {{EMNLP}} 2020},
  author = {Feng, Zhangyin and Guo, Daya and Tang, Duyu and Duan, Nan and Feng, Xiaocheng and Gong, Ming and Shou, Linjun and Qin, Bing and Liu, Ting and Jiang, Daxin and Zhou, Ming},
  year = {2020},
  month = nov,
  pages = {1536--1547},
  publisher = {{Association for Computational Linguistics}},
  address = {{Online}},
  doi = {10.18653/v1/2020.findings-emnlp.139}
}

@inproceedings{Lin2020Bridging,
  title = {Bridging {{Textual}} and {{Tabular Data}} for {{Cross-Domain Text-to-SQL Semantic Parsing}}},
  booktitle = {Findings of the {{Association}} for {{Computational Linguistics}}: {{EMNLP}} 2020},
  author = {Lin, Xi Victoria and Socher, Richard and Xiong, Caiming},
  year = {2020},
  month = nov,
  pages = {4870--4888},
  publisher = {{Association for Computational Linguistics}},
  address = {{Online}},
  doi = {10.18653/v1/2020.findings-emnlp.438}
}

@inproceedings{Wang2020RATSQLa,
  title = {{{RAT-SQL}}: {{Relation-Aware Schema Encoding}} and {{Linking}} for {{Text-to-SQL Parsers}}},
  shorttitle = {{{RAT-SQL}}},
  booktitle = {Proceedings of the 58th {{Annual Meeting}} of the {{Association}} for {{Computational Linguistics}}},
  author = {Wang, Bailin and Shin, Richard and Liu, Xiaodong and Polozov, Oleksandr and Richardson, Matthew},
  year = {2020},
  month = jul,
  pages = {7567--7578},
  publisher = {{Association for Computational Linguistics}},
  address = {{Online}},
  doi = {10.18653/v1/2020.acl-main.677}
}

@inproceedings{horn1990hadamard,
  title={The hadamard product},
  author={Horn, Roger A},
  booktitle={Proc. symp. appl. math},
  volume={40},
  pages={87--169},
  year={1990}
}

@book{horn2012matrix,
  title={Matrix analysis},
  author={Horn, Roger A and Johnson, Charles R},
  year={2012},
  publisher={Cambridge university press}
}

@article{li2024can,
  title={Can llm already serve as a database interface? a big bench for large-scale database grounded text-to-sqls},
  author={Li, Jinyang and Hui, Binyuan and Qu, Ge and Yang, Jiaxi and Li, Binhua and Li, Bowen and Wang, Bailin and Qin, Bowen and Geng, Ruiying and Huo, Nan and others},
  journal={Advances in Neural Information Processing Systems},
  volume={36},
  year={2024}
}

@inproceedings{yu-etal-2018-spider,
    title = "{S}pider: A Large-Scale Human-Labeled Dataset for Complex and Cross-Domain Semantic Parsing and Text-to-{SQL} Task",
    author = "Yu, Tao  and
      Zhang, Rui  and
      Yang, Kai  and
      Yasunaga, Michihiro  and
      Wang, Dongxu  and
      Li, Zifan  and
      Ma, James  and
      Li, Irene  and
      Yao, Qingning  and
      Roman, Shanelle  and
      Zhang, Zilin  and
      Radev, Dragomir",
    editor = "Riloff, Ellen  and
      Chiang, David  and
      Hockenmaier, Julia  and
      Tsujii, Jun{'}ichi",
    booktitle = "Proceedings of the 2018 Conference on Empirical Methods in Natural Language Processing",
    month = oct # "-" # nov,
    year = "2018",
    address = "Brussels, Belgium",
    publisher = "Association for Computational Linguistics",
    url = "https://aclanthology.org/D18-1425/",
    doi = "10.18653/v1/D18-1425",
    pages = "3911--3921"
}

@article{lee2022ehrsql,
  title={Ehrsql: A practical text-to-sql benchmark for electronic health records},
  author={Lee, Gyubok and Hwang, Hyeonji and Bae, Seongsu and Kwon, Yeonsu and Shin, Woncheol and Yang, Seongjun and Seo, Minjoon and Kim, Jong-Yeup and Choi, Edward},
  journal={Advances in Neural Information Processing Systems},
  volume={35},
  pages={15589--15601},
  year={2022}
}

@article{saparina2024ambrosia,
  title={Ambrosia: A benchmark for parsing ambiguous questions into database queries},
  author={Saparina, Irina and Lapata, Mirella},
  journal={Advances in Neural Information Processing Systems},
  volume={37},
  pages={90600--90628},
  year={2024}
}

@inproceedings{leispider,
      title={Spider 2.0: Evaluating Language Models on Real-World Enterprise Text-to-SQL Workflows}, 
      author={Fangyu Lei and Jixuan Chen and Yuxiao Ye and Ruisheng Cao and Dongchan Shin and Hongjin Su and Zhaoqing Suo and Hongcheng Gao and Wenjing Hu and Pengcheng Yin and Victor Zhong and Caiming Xiong and Ruoxi Sun and Qian Liu and Sida Wang and Tao Yu},
      year={2024},
      eprint={2411.07763},
      archivePrefix={arXiv},
      primaryClass={cs.CL},
      url={https://arxiv.org/abs/2411.07763}, 
}

@article{johnson2023mimic,
  title={MIMIC-IV, a freely accessible electronic health record dataset},
  author={Johnson, Alistair EW and Bulgarelli, Lucas and Shen, Lu and Gayles, Alvin and Shammout, Ayad and Horng, Steven and Pollard, Tom J and Hao, Sicheng and Moody, Benjamin and Gow, Brian and others},
  journal={Scientific data},
  volume={10},
  number={1},
  pages={1},
  year={2023},
  publisher={Nature Publishing Group UK London}
}

@article{lee2024overview,
  title={Overview of the ehrsql 2024 shared task on reliable text-to-sql modeling on electronic health records},
  author={Lee, Gyubok and Kweon, Sunjun and Bae, Seongsu and Choi, Edward},
  journal={arXiv preprint arXiv:2405.06673},
  year={2024}
}

@misc{Arcadinho2022T5QL,
  title = {{{T5QL}}: {{Taming}} Language Models for {{SQL}} Generation},
  shorttitle = {{{T5QL}}},
  author = {Arcadinho, Samuel and Aparício, David and Veiga, Hugo and Alegria, António},
  date = {2022-09-21},
  year = {2022},
  number = {arXiv:2209.10254},
  eprint = {2209.10254},
  eprinttype = {arxiv},
  primaryclass = {cs},
  publisher = {{arXiv}},
  url = {http://arxiv.org/abs/2209.10254},
  urldate = {2022-12-11},
  archiveprefix = {arXiv}
}

@inproceedings{Yu2021Grappa,
  title={GraPPa: Grammar-Augmented Pre-Training for Table Semantic Parsing},
  author={Tao Yu and Chien-Sheng Wu and Xi Victoria Lin and Bailin Wang and Yi Chern Tan and Xinyi Yang and Dragomir Radev and Richard Socher and Caiming Xiong},
  booktitle={International Conference on Learning Representations},
  year={2021},
  url={https://arxiv.org/abs/2009.13845}
}

@inproceedings{li2022resdsql,
  author = {Haoyang Li and Jing Zhang and Cuiping Li and Hong Chen},
  title = "RESDSQL: Decoupling Schema Linking and Skeleton Parsing for Text-to-SQL",
  booktitle = "AAAI",
  year = "2023"
}

@inproceedings{chen-etal-2023-sqledit,
    title = "Text-to-{SQL} Error Correction with Language Models of Code",
    author = "Chen, Ziru  and
      Chen, Shijie  and
      White, Michael  and
      Mooney, Raymond  and
      Payani, Ali  and
      Srinivasa, Jayanth  and
      Su, Yu  and
      Sun, Huan",
    booktitle = "Proceedings of the 61st Annual Meeting of the Association for Computational Linguistics (Volume 2: Short Papers)",
    month = jul,
    year = "2023",
    address = "Toronto, Canada",
    publisher = "Association for Computational Linguistics",
    url = "https://aclanthology.org/2023.acl-short.117",
    doi = "10.18653/v1/2023.acl-short.117",
    pages = "1359--1372",
}

@ARTICLE{bugfixhard1,
  author={Jorgensen, Magne and Shepperd, Martin},
  journal={IEEE Transactions on Software Engineering}, 
  title={A Systematic Review of Software Development Cost Estimation Studies}, 
  year={2007},
  volume={33},
  number={1},
  pages={33-53},
  doi={10.1109/TSE.2007.256943}}

@article{wei2022chain,
  title={Chain-of-thought prompting elicits reasoning in large language models},
  author={Wei, Jason and Wang, Xuezhi and Schuurmans, Dale and Bosma, Maarten and Xia, Fei and Chi, Ed and Le, Quoc V and Zhou, Denny and others},
  journal={Advances in neural information processing systems},
  volume={35},
  pages={24824--24837},
  year={2022}
}

@inproceedings{chen2023error,
  title={Error Detection for Text-to-SQL Semantic Parsing},
  author={Chen, Shijie and Chen, Ziru and Sun, Huan and Su, Yu},
  booktitle={EMNLP (Findings)},
  year={2023}
}

@inproceedings{kim2023sure,
  title={Sure: Improving open-domain question answering of llms via summarized retrieval},
  author={Kim, Jaehyung and Nam, Jaehyun and Mo, Sangwoo and Park, Jongjin and Lee, Sang-Woo and Seo, Minjoon and Ha, Jung-Woo and Shin, Jinwoo},
  booktitle={The Twelfth International Conference on Learning Representations},
  year={2023}
}

@inproceedings{dhuliawala2024chain,
  title={Chain-of-Verification Reduces Hallucination in Large Language Models},
  author={Dhuliawala, Shehzaad and Komeili, Mojtaba and Xu, Jing and Raileanu, Roberta and Li, Xian and Celikyilmaz, Asli and Weston, Jason},
  booktitle={Findings of the Association for Computational Linguistics ACL 2024},
  pages={3563--3578},
  year={2024}
}

@INPROCEEDINGS{bugfixhard2,
  author={Weiss, Cathrin and Premraj, Rahul and Zimmermann, Thomas and Zeller, Andreas},
  booktitle={Fourth International Workshop on Mining Software Repositories (MSR'07:ICSE Workshops 2007)}, 
  title={How Long Will It Take to Fix This Bug?}, 
  year={2007},
  volume={},
  number={},
  pages={1-1},
  doi={10.1109/MSR.2007.13}}

@article{10.14778/3681954.3682003,
    author = {Li, Boyan and Luo, Yuyu and Chai, Chengliang and Li, Guoliang and Tang, Nan},
    title = {The Dawn of Natural Language to SQL: Are We Fully Ready?},
    year = {2024},
    issue_date = {July 2024},
    publisher = {VLDB Endowment},
    volume = {17},
    number = {11},
    issn = {2150-8097},
    url = {https://doi.org/10.14778/3681954.3682003},
    doi = {10.14778/3681954.3682003},
    journal = {Proc. VLDB Endow.},
    month = jul,
    pages = {3318–3331},
    numpages = {14}
}

@article{DBLP:journals/corr/abs-2204-00498,
	author       = {Nitarshan Rajkumar and et al.},
	title        = {Evaluating the Text-to-SQL Capabilities of Large Language Models},
	journal      = {CoRR},
	year         = {2022}
}

@inproceedings{10.1145/2588555.2594519,
    author = {Li, Fei and Jagadish, Hosagrahar V},
    title = {NaLIR: an interactive natural language interface for querying relational databases},
    year = {2014},
    booktitle = {ACM SIGMOD}
}

@inproceedings{DBLP:conf/sigmod/Katsogiannis-Meimarakis21,
  title={A deep dive into deep learning approaches for text-to-sql systems},
  author={Katsogiannis-Meimarakis, George and Koutrika, Georgia},
  booktitle={SIGMOD},
  year={2021}
}

@inproceedings{xiao2016sequence,
    title={Sequence-based structured prediction for semantic parsing},
    author={Xiao, Chunyang and Dymetman, Marc and Gardent, Claire},
    booktitle={ACL},
    year={2016}
}

@article{lin2019grammar,
  title={Grammar-based neural text-to-sql generation},
  author={Lin, Kevin and Bogin, Ben and Neumann, Mark and et al.},
  journal={arXiv:1905.13326},
  year={2019}
}

@article{bogin2019representing,
  title={Representing schema structure with graph neural networks for text-to-SQL parsing},
  author={Bogin, Ben and Gardner, Matt and et al.},
  journal={arXiv:1905.06241},
  year={2019}
}

@article{li2023graphix,
  title={Graphix-t5: Mixing pre-trained transformers with graph-aware layers for text-to-sql parsing},
  author={Li, Jinyang and et al.},
  journal={arXiv:2301.07507},
  year={2023}
}

@article{bazaga2023sqlformer,
  title={SQLformer: Deep Auto-Regressive Query Graph Generation for Text-to-SQL Translation},
  author={Bazaga, Adri{\'a}n and Li{\`o}, Pietro and et al.},
  journal={arXiv:2310.18376},
  year={2023}
}

@article{gu2023interleaving,
  title={Interleaving Pre-Trained Language Models and Large Language Models for Zero-Shot NL2SQL Generation.},
  author={Gu, Zihui and Fan, Ju and Tang, Nan and et al.},
  journal={arXiv:2306.08891},
  year={2023}
}

@inproceedings{rai-etal-2023-improving,
  title={Improving Generalization in Language Model-based Text-to-SQL Semantic Parsing: Two Simple Semantic Boundary-based Techniques},
  author={Rai, Daking and Wang, Bailin and Zhou, Yilun and et al.},
  booktitle={ACL},
  year={2023}
}

@article{scholak2021picard,
  title={PICARD: Parsing incrementally for constrained auto-regressive decoding from language models},
  author={Scholak, Torsten and Schucher, Nathan and Bahdanau, Dzmitry},
  journal={arXiv preprint arXiv:2109.05093},
  year={2021}
}

@article{qi2022rasat,
  title={Rasat: Integrating relational structures into pretrained seq2seq model for text-to-sql},
  author={Qi, Jiexing and Tang, Jingyao and He, Ziwei and et al.},
  journal={arXiv:2205.06983},
  year={2022}
}

@inproceedings{hui2022s,
  title={S2SQL: Injecting Syntax to Question-Schema Interaction Graph Encoder for Text-to-SQL Parsers},
  author={Hui, Binyuan and et al.},
  booktitle={Findings of ACL},
  year={2022}
}

@article{gong2025sqlens,
  title={SQLens: An End-to-End Framework for Error Detection and Correction in Text-to-SQL},
  author={Gong, Yue and Lei, Chuan and Qin, Xiao and Vaidya, Kapil and Narayanaswamy, Balakrishnan and Kraska, Tim},
  journal={Advances in Neural Information Processing Systems},
  year={2025}
}

@inproceedings{hu2023importance,
  title={Importance of synthesizing high-quality data for text-to-SQL parsing},
  author={Hu, Yiqun and Zhao, Yiyun and Jiang, Jiarong and Lan, Wuwei and Zhu, Henghui and Chauhan, Anuj and Li, Alexander Hanbo and Pan, Lin and Wang, Jun and Hang, Chung-Wei and others},
  booktitle={Findings of the Association for Computational Linguistics: ACL 2023},
  pages={1327--1343},
  year={2023}
}

@inproceedings{tarbell2024towards,
  title={Towards understanding the generalization of medical text-to-sql models and datasets},
  author={Tarbell, Richard and Choo, Kim-Kwang Raymond and Dietrich, Glenn and Rios, Anthony},
  booktitle={AMIA Annual Symposium Proceedings},
  volume={2023},
  pages={669},
  year={2024}
}

@article{ScienceBenchmark,
author = {Zhang, Yi and Deriu, Jan and Katsogiannis-Meimarakis, George and Kosten, Catherine and Koutrika, Georgia and Stockinger, Kurt},
title = {ScienceBenchmark: A Complex Real-World Benchmark for Evaluating Natural Language to SQL Systems},
year = {2023},
issue_date = {December 2023},
publisher = {VLDB Endowment},
volume = {17},
number = {4},
issn = {2150-8097},
url = {https://doi.org/10.14778/3636218.3636225},
doi = {10.14778/3636218.3636225},
journal = {Proc. VLDB Endow.},
month = dec,
pages = {685–698},
numpages = {14}
}

@inproceedings{lee-etal-2025-mcs,
    title = "{MCS}-{SQL}: Leveraging Multiple Prompts and Multiple-Choice Selection For Text-to-{SQL} Generation",
    author = "Lee, Dongjun  and
      Park, Choongwon  and
      Kim, Jaehyuk  and
      Park, Heesoo",
    editor = "Rambow, Owen  and
      Wanner, Leo  and
      Apidianaki, Marianna  and
      Al-Khalifa, Hend  and
      Eugenio, Barbara Di  and
      Schockaert, Steven",
    booktitle = "Proceedings of the 31st International Conference on Computational Linguistics",
    month = jan,
    year = "2025",
    address = "Abu Dhabi, UAE",
    publisher = "Association for Computational Linguistics",
    url = "https://aclanthology.org/2025.coling-main.24/",
    pages = "337--353"
}

@article{li2024codes,
  title={Codes: Towards building open-source language models for text-to-sql},
  author={Li, Haoyang and Zhang, Jing and Liu, Hanbing and Fan, Ju and Zhang, Xiaokang and Zhu, Jun and Wei, Renjie and Pan, Hongyan and Li, Cuiping and Chen, Hong},
  journal={Proceedings of the ACM on Management of Data},
  volume={2},
  number={3},
  pages={1--28},
  year={2024},
  publisher={ACM New York, NY, USA}
}

@inproceedings{duan2025dsqg,
  title={DSQG-Syn: Synthesizing High-quality Data for Text-to-SQL Parsing by Domain Specific Question Generation},
  author={Duan, Shaoming and Wu, Youxuan and Liu, Chuanyi and Zhang, Yuhao and Wang, Zirui and Han, Peiyi and Yu, Shengyuan and Yan, Liang and Liang, Yingwei},
  booktitle={Findings of the Association for Computational Linguistics: NAACL 2025},
  pages={2971--2989},
  year={2025}
}

@article{OmniSQL,
author = {Li, Haoyang and Wu, Shang and Zhang, Xiaokang and Huang, Xinmei and Zhang, Jing and Jiang, Fuxin and Wang, Shuai and Zhang, Tieying and Chen, Jianjun and Shi, Rui and Chen, Hong and Li, Cuiping},
title = {OmniSQL: Synthesizing High-Quality Text-to-SQL Data at Scale},
year = {2025},
issue_date = {July 2025},
publisher = {VLDB Endowment},
volume = {18},
number = {11},
issn = {2150-8097},
url = {https://doi.org/10.14778/3749646.3749723},
doi = {10.14778/3749646.3749723},
journal = {Proc. VLDB Endow.},
month = sep,
pages = {4695–4709},
numpages = {15}
}

@inproceedings{yang-etal-2024-synthesizing,
    title = "Synthesizing Text-to-{SQL} Data from Weak and Strong {LLM}s",
    author = "Yang, Jiaxi  and
      Hui, Binyuan  and
      Yang, Min  and
      Yang, Jian  and
      Lin, Junyang  and
      Zhou, Chang",
    editor = "Ku, Lun-Wei  and
      Martins, Andre  and
      Srikumar, Vivek",
    booktitle = "Proceedings of the 62nd Annual Meeting of the Association for Computational Linguistics (Volume 1: Long Papers)",
    month = aug,
    year = "2024",
    address = "Bangkok, Thailand",
    publisher = "Association for Computational Linguistics",
    url = "https://aclanthology.org/2024.acl-long.425/",
    doi = "10.18653/v1/2024.acl-long.425",
    pages = "7864--7875"
}

@INPROCEEDINGS{10484107,
  author={Shen, Hao and Shen, Ran and Sun, Gang and Li, Yiling and Wang, Yifan and Zhang, Pengcheng},
  booktitle={2023 International Conference on Algorithms, Computing and Data Processing (ACDP)}, 
  title={Sequential Feature Augmentation for Robust Text-to-SQL}, 
  year={2023},
  volume={},
  number={},
  pages={217-223},
  doi={10.1109/ACDP59959.2023.00042}
}

@inproceedings{dao2022flashattention,
  title={Flash{A}ttention: Fast and Memory-Efficient Exact Attention with {IO}-Awareness},
  author={Dao, Tri and Fu, Daniel Y. and Ermon, Stefano and Rudra, Atri and R{\'e}, Christopher},
  booktitle={Advances in Neural Information Processing Systems (NeurIPS)},
  year={2022}
}

@inproceedings{dao2023flashattention2,
  title={Flash{A}ttention-2: Faster Attention with Better Parallelism and Work Partitioning},
  author={Dao, Tri},
  booktitle={International Conference on Learning Representations (ICLR)},
  year={2024}
}

@article{agarap2018deep,
  title={Deep learning using rectified linear units (relu)},
  author={Agarap, Abien Fred},
  journal={arXiv preprint arXiv:1803.08375},
  year={2018}
}

@article{loshchilov2017decoupled,
  title={Decoupled weight decay regularization},
  author={Loshchilov, Ilya and Hutter, Frank},
  journal={arXiv preprint arXiv:1711.05101},
  year={2017}
}

@article{li2023towards,
  title={Towards general text embeddings with multi-stage contrastive learning},
  author={Li, Zehan and Zhang, Xin and Zhang, Yanzhao and Long, Dingkun and Xie, Pengjun and Zhang, Meishan},
  journal={arXiv preprint arXiv:2308.03281},
  year={2023}
}

@inproceedings{xiao2024c,
  title={C-pack: Packed resources for general chinese embeddings},
  author={Xiao, Shitao and Liu, Zheng and Zhang, Peitian and Muennighoff, Niklas and Lian, Defu and Nie, Jian-Yun},
  booktitle={Proceedings of the 47th international ACM SIGIR conference on research and development in information retrieval},
  pages={641--649},
  year={2024}
}

@inproceedings{shi2023lmc,
    title={{LMC}: Fast Training of {GNN}s via Subgraph Sampling with Provable Convergence},
    author={Zhihao Shi and Xize Liang and Jie Wang},
    booktitle={International Conference on Learning Representations},
    year={2023},
    url={https://openreview.net/forum?id=5VBBA91N6n}
}

@article{ding2022sketch,
  title={Sketch-GNN: Scalable graph neural networks with sublinear training complexity},
  author={Ding, Mucong and Rabbani, Tahseen and An, Bang and Wang, Evan and Huang, Furong},
  journal={Advances in neural information processing systems},
  volume={35},
  pages={2930--2943},
  year={2022}
}

@article{xue2024haste,
  title={Haste makes waste: A simple approach for scaling graph neural networks},
  author={Xue, Rui and Zhao, Tong and Shah, Neil and Liu, Xiaorui},
  journal={International Conference on Machine Learning},
  year={2025}
}

@article{dong2023megraph,
  title={MeGraph: capturing long-range interactions by alternating local and hierarchical aggregation on multi-scaled graph hierarchy},
  author={Dong, Honghua and Xu, Jiawei and Yang, Yu and Zhao, Rui and Wu, Shiwen and Yuan, Chun and Li, Xiu and Maddison, Chris J and Han, Lei},
  journal={Advances in Neural Information Processing Systems},
  volume={36},
  pages={63609--63641},
  year={2023}
}

@inproceedings{park2022vldb,
 author    = {Yeonhong Park and Sunhong Min and Jae W. Lee},
 title     = {Ginex: SSD-enabled Billion-scale Graph Neural Network Training on a Single Machine via Provably Optimal In-memory Caching},
 booktitle = {Proceedings of the VLDB Endowment},
 volume    = {15},
 number    = {11},
 year      = {2022}
}

@article{robinson2020contrastive,
  title={Contrastive learning with hard negative samples},
  author={Robinson, Joshua and Chuang, Ching-Yao and Sra, Suvrit and Jegelka, Stefanie},
  journal={arXiv preprint arXiv:2010.04592},
  year={2020}
}

@article{gao2021simcse,
  title={Simcse: Simple contrastive learning of sentence embeddings},
  author={Gao, Tianyu and Yao, Xingcheng and Chen, Danqi},
  journal={arXiv preprint arXiv:2104.08821},
  year={2021}
}

@inproceedings{wu2022esimcse,
  title={Esimcse: Enhanced sample building method for contrastive learning of unsupervised sentence embedding},
  author={Wu, Xing and Gao, Chaochen and Zang, Liangjun and Han, Jizhong and Wang, Zhongyuan and Hu, Songlin},
  booktitle={Proceedings of the 29th international conference on computational linguistics},
  pages={3898--3907},
  year={2022}
}

@article{chawla2002smote,
  title={SMOTE: synthetic minority over-sampling technique},
  author={Chawla, Nitesh V and Bowyer, Kevin W and Hall, Lawrence O and Kegelmeyer, W Philip},
  journal={Journal of artificial intelligence research},
  volume={16},
  pages={321--357},
  year={2002}
}
